\pdfoutput=1

\documentclass[11pt]{article}

\usepackage[final]{ACL2023}

\usepackage{times}
\usepackage{latexsym}

\usepackage[T1]{fontenc}

\usepackage[utf8]{inputenc}

\usepackage{microtype}
\usepackage{subfigure}
\usepackage{tabularx}
\newcolumntype{C}{>{\centering\arraybackslash}X}

\usepackage{inconsolata}

\usepackage{booktabs}
\usepackage{graphicx}

\title{RAG+: Enhancing Retrieval-Augmented Generation with Application-Aware Reasoning}

\author {
    Yu Wang\textsuperscript{†$\heartsuit$},
    Shiwan Zhao\textsuperscript{$\triangle$},
    Zhihu Wang\textsuperscript{$\diamondsuit$},
    Ming Fan\textsuperscript{*$\heartsuit$},
    Xicheng Zhang\textsuperscript{$\heartsuit$},\\
    \textbf{Yubo Zhang}\textsuperscript{$\diamondsuit$},
    \textbf{Zhengfan Wang}\textsuperscript{$\heartsuit$},
    \textbf{Heyuan Huang}\textsuperscript{$\diamondsuit$},
    \textbf{Ting Liu}\textsuperscript{$\heartsuit$}\\
    \textsuperscript{$\heartsuit$}Xi’an Jiaotong University\quad
    \textsuperscript{$\diamondsuit$}Huawei Technologies Ltd.\quad
    \textsuperscript{$\triangle$}Nankai University\\
    \texttt{\{uyleewang,xichengzhang,wangzf7241\}@stu.xjtu.edu.cn}\\
    \texttt{\{mingfan,tingliu\}@mail.xjtu.edu.cn} \quad
    \texttt{zhaosw@gmail.com}\\
    \texttt{\{wangzhihu3,zhangyubo20,huangheyuan\}@huawei.com}
}

\begin{document}
\maketitle
\renewcommand{\thefootnote}{}
\footnotetext{\textsuperscript{†}Work done during an internship at Huawei Technologies Ltd.}
\footnotetext{\textsuperscript{*}Corresponding Author}
\renewcommand{\thefootnote}{\arabic{footnote}}
\begin{abstract}
The integration of external knowledge through Retrieval-Augmented Generation (RAG) has become foundational in enhancing large language models (LLMs) for knowledge-intensive tasks. However, existing RAG paradigms often overlook the cognitive step of applying knowledge, leaving a gap between retrieved facts and task-specific reasoning.
In this work, we introduce \textbf{RAG+}, a principled and modular extension that explicitly incorporates application-aware reasoning into the RAG pipeline.
RAG+ constructs a dual corpus consisting of knowledge and aligned application examples, created either manually or automatically, and retrieves both jointly during inference. This design enables LLMs not only to access relevant information but also to apply it within structured, goal-oriented reasoning processes.
Experiments across mathematical, legal, and medical domains, conducted on multiple models, demonstrate that RAG+ consistently outperforms standard RAG variants, achieving average improvements of 3–5\%, and peak gains up to 13.5\% in complex scenarios. 
By bridging retrieval with actionable application, RAG+ advances a more cognitively grounded framework for knowledge integration, representing a step toward more interpretable and capable LLMs.
\end{abstract}

\section{Introduction}

Large language models (LLMs) have demonstrated strong performance across a broad range of natural language processing tasks~\citep{li2025begonia, tang2025aspect}. To further improve their capabilities, Retrieval-Augmented Generation (RAG) has emerged as a widely adopted framework.
By equipping LLMs with access to external knowledge sources, RAG enables the dynamic retrieval of up-to-date information at inference time, leading to significant gains in knowledge-intensive scenarios~\citep{li2025ragSurvey, mostafa2025ragIntent}.

However, existing RAG methods often focus on lexical or semantic similarity when retrieving knowledge, paying little attention to how the retrieved content should be applied in downstream tasks. 
\begin{figure}[t]
\centering
\includegraphics[width=0.497\textwidth]{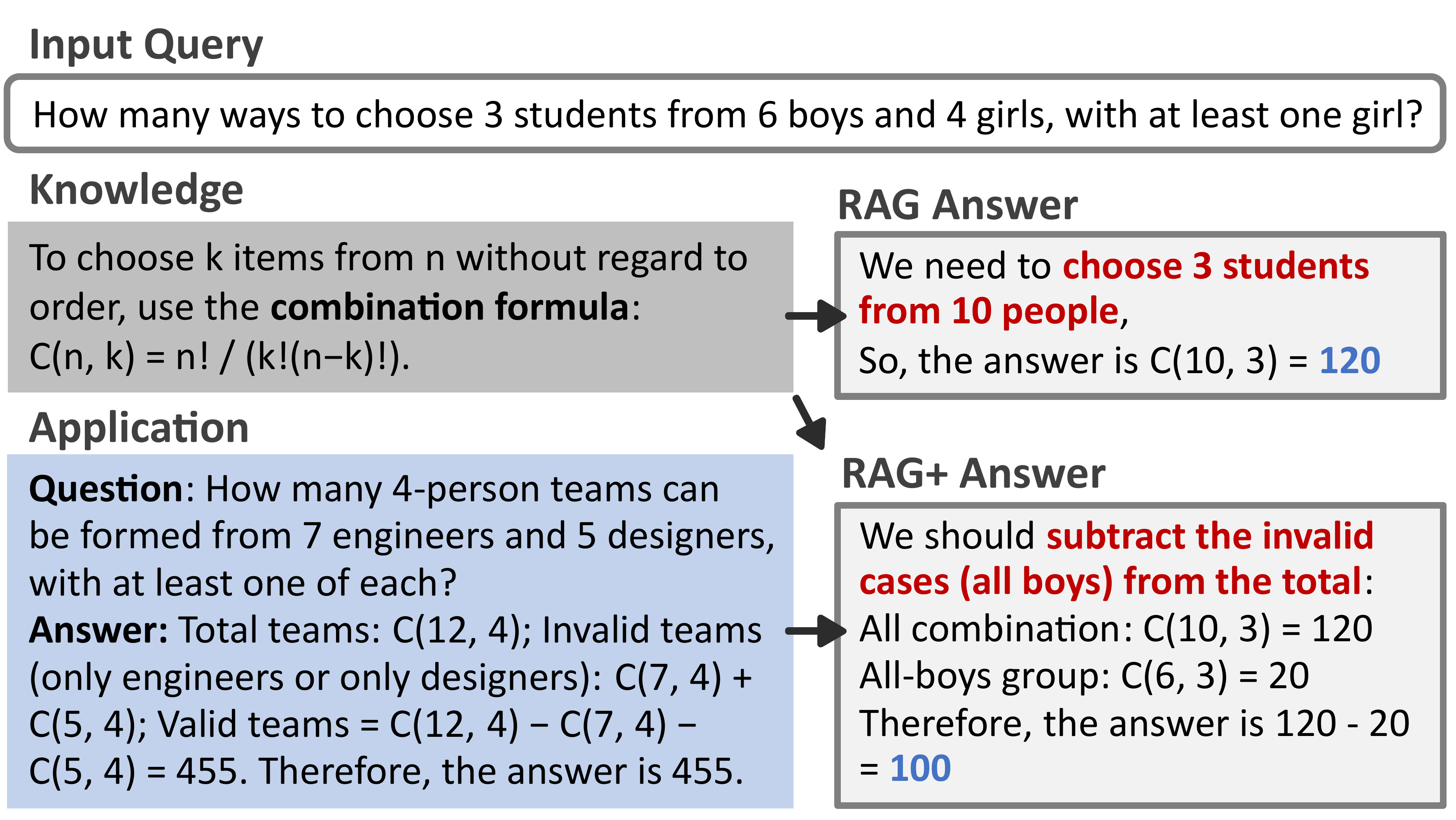}
\caption{An Illustrative Case of RAG and RAG+: Knowledge Only vs. Knowledge with Application.}
\label{fig:casemain}
\end{figure}
While effective in factual recall and open-domain question answering, RAG frequently underperforms on domain-specific reasoning tasks~\citep{lin2025cancerRAG, hayashi2025iterkey, ammann2025securingRAG}, where solving complex problems requires not only relevant information but also reasoning about how to use it to arrive at a solution. Figure~\ref{fig:casemain} illustrates a representative example.

RA-DIT~\cite{lin2023ra} fine-tunes both the retriever and generator in a dual-instruction manner, aligning retrieval more closely with what the model needs to generate accurate responses.
Other recent RAG extensions address the gap between retrieval and application by decomposing reasoning tasks into smaller steps and retrieving relevant knowledge for each substep~\citep{singh2025agenticretrievalaugmentedgenerationsurvey, xiong2025raggym, wang2024rat}.
However, they provide limited guidance on how to apply retrieved knowledge. This shortcoming hampers performance in procedural reasoning tasks that require understanding both the solution process and the associated underlying concepts. 

This limitation reflects insights from educational psychology. Bloom’s Taxonomy identifies “applying” knowledge as a distinct cognitive skill that goes beyond simple recall~\citep{bloom2010taxonomy}. Similarly, cognitive architectures like ACT-R distinguish between declarative memory (facts) and procedural memory (skills), suggesting that coupling factual knowledge with procedural examples enhances performance on complex tasks~\citep{anderson2004integrated}. Consistently, Re-TASK~\cite{wang2024re} introduces the concept of capability items, emphasizing that accomplishing domain-specific tasks requires jointly leveraging domain knowledge and task-specific skills. 

Motivated by these insights, we propose RAG+, a simple yet effective extension to the RAG framework that enhances reasoning by bridging retrieval and generation with an application-aware stage. 
Instead of retrieving only relevant knowledge, RAG+ additionally retrieves examples that demonstrate how the knowledge is applied in practice-such as structured reasoning chains and stepwise solutions-to ground the model’s output in task-relevant usage and to improve reasoning accuracy.

RAG+ builds a dual corpus: one stores domain knowledge, and the other contains automatically generated application instances aligned to each fact. At inference time, the system retrieves relevant knowledge based on the input query and then fetches aligned application examples to provide practical context. This design encourages the model not only to recall factual content but also to produce outputs that follow grounded reasoning patterns based on prior usage.
RAG+ is modular and retrieval-agnostic; it can be integrated into any existing RAG pipeline without changing the model architecture or requiring additional fine-tuning.

We evaluate RAG+ across three reasoning-intensive domains: mathematics, medicine, and legal, using four representative RAG variants: vanilla RAG, Answer-First RAG, GraphRAG, and Rerank RAG.
Experiments show that RAG+ consistently improves performance. Notably, Qwen2.5-72B improves from 76.5\% to 87.5\% on legal prediction, and LLaMA3.3-70B rises from 78.2\% to 86.5\% on medical QA. Even smaller models like DS-Qwen-7B benefit, demonstrating the broad effectiveness of application-aware augmentation. These results demonstrate that bridging retrieval and reasoning through application-aware augmentation can yield substantial gains, especially in domains where accurate reasoning is essential.

Our contributions are as follows:
\begin{itemize}
    \item We identify a key limitation of existing RAG systems: the absence of an application-aware step, which impairs reasoning in complex tasks.
    \item We introduce RAG+, a modular and plug-and-play extension that jointly retrieves factual knowledge and its aligned usage examples to better support downstream reasoning.
    \item We validate the effectiveness of RAG+ through comprehensive experiments, demonstrating consistent gains across multiple domains and retrieval strategies.
\end{itemize}

\section{Related Work}

Retrieval-Augmented Generation (RAG) has become a widely adopted framework for enhancing large language models (LLMs), especially in knowledge-intensive tasks. A standard RAG pipeline includes query formulation, corpus construction, retrieval, and answer generation. Prior work has sought improvements at each stage.

For retrieval quality, Rewrite-Retrieve-Read~\citep{ma-etal-2023-query} rewrites queries to better align with document style. $R^2AG$~\citep{ye-etal-2024-r2ag} uses a retrieval-aware encoder to highlight key signals, while Query2Doc~\citep{wang2023query2doc} expands queries into pseudo-documents to clarify intent. In terms of corpus construction, GraphRAG~\citep{edge2024local} integrates knowledge graphs to support entity-centric reasoning. To enhance retrieval precision, Reranking~\citep{wang-etal-2024-searching} and filtering techniques~\citep{wang2023learning, pickett2025betterragusingrelevant} further refine retrieved content. 

Beyond these enhancements, recent work has tackled more complex reasoning tasks through modular and agent-based RAG variants. Agentic RAG~\citep{singh2025agenticretrievalaugmentedgenerationsurvey} and RAG-Gym~\citep{xiong2025raggym} decompose tasks into subtasks handled by specialized agents or workflows. OPEN-RAG~\citep{islam-etal-2024-open} introduces agent-based decomposition and context selection, while RAT~\citep{wang2024rat} uses Chain-of-Thought prompting to support step-wise retrieval.

Despite these efforts, most RAG methods remain optimized for fact-centric tasks such as open-domain QA. In reasoning-centric domains such as mathematics, retrieving the right facts is only the first step. The model must also understand how to apply them to reach a specific goal. Our work addresses this gap by introducing an application-aware step that explicitly guides how retrieved knowledge is used.

\section{Methods}
\begin{figure*}[t]
  \centering
   \subfigure[Construction Stage: Building an Application Corpus Aligned with the Knowledge Corpus.]{
    \label{fig:construction}
    \includegraphics[width=0.3933\textwidth]{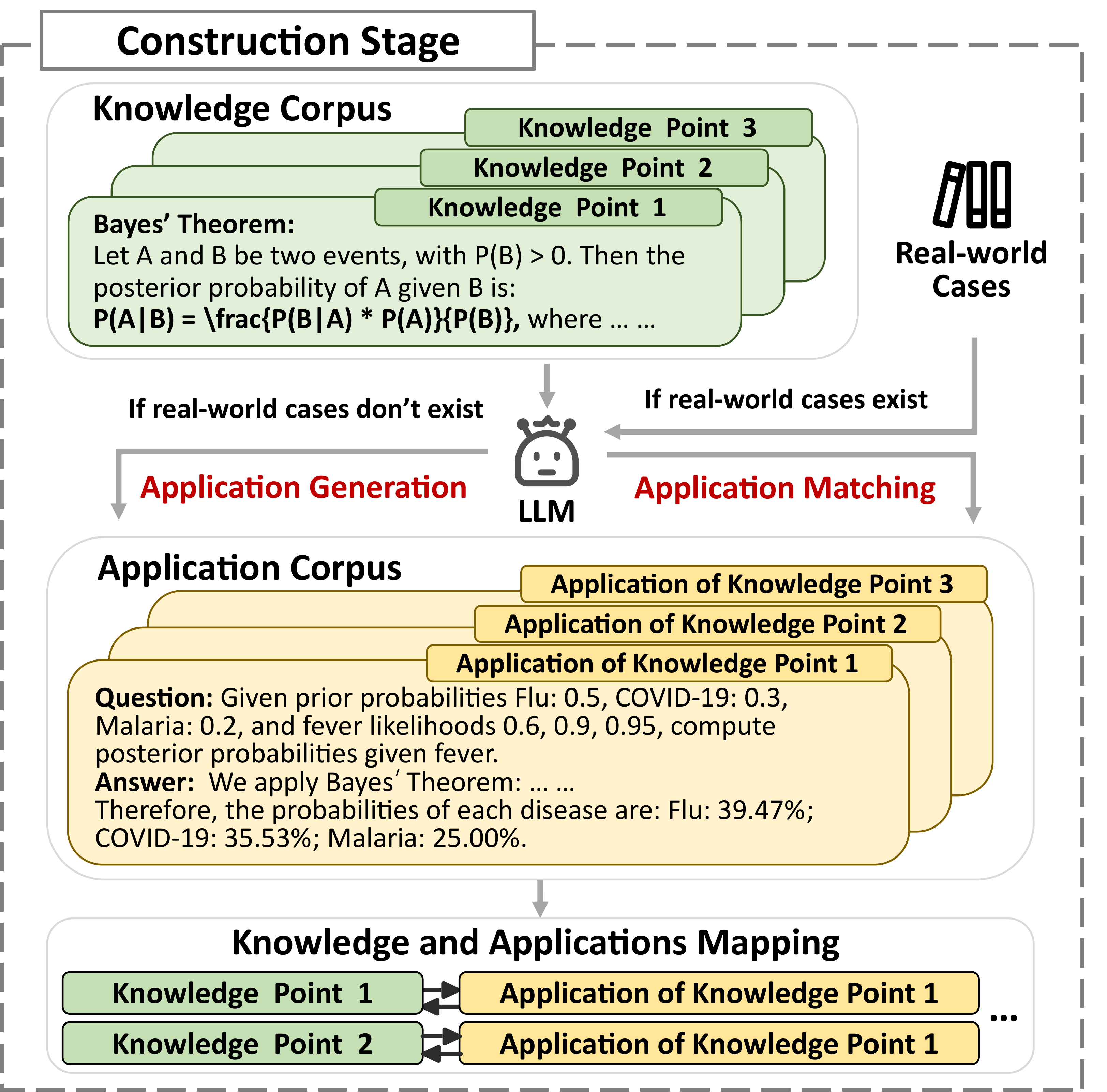}} 
    \hspace{0.05cm}
  \subfigure[Inference Stage: Retrieving Aligned Knowledge and Application Examples to Generate Output.]{
    \label{fig:inference}
\includegraphics[width=0.5802\textwidth]{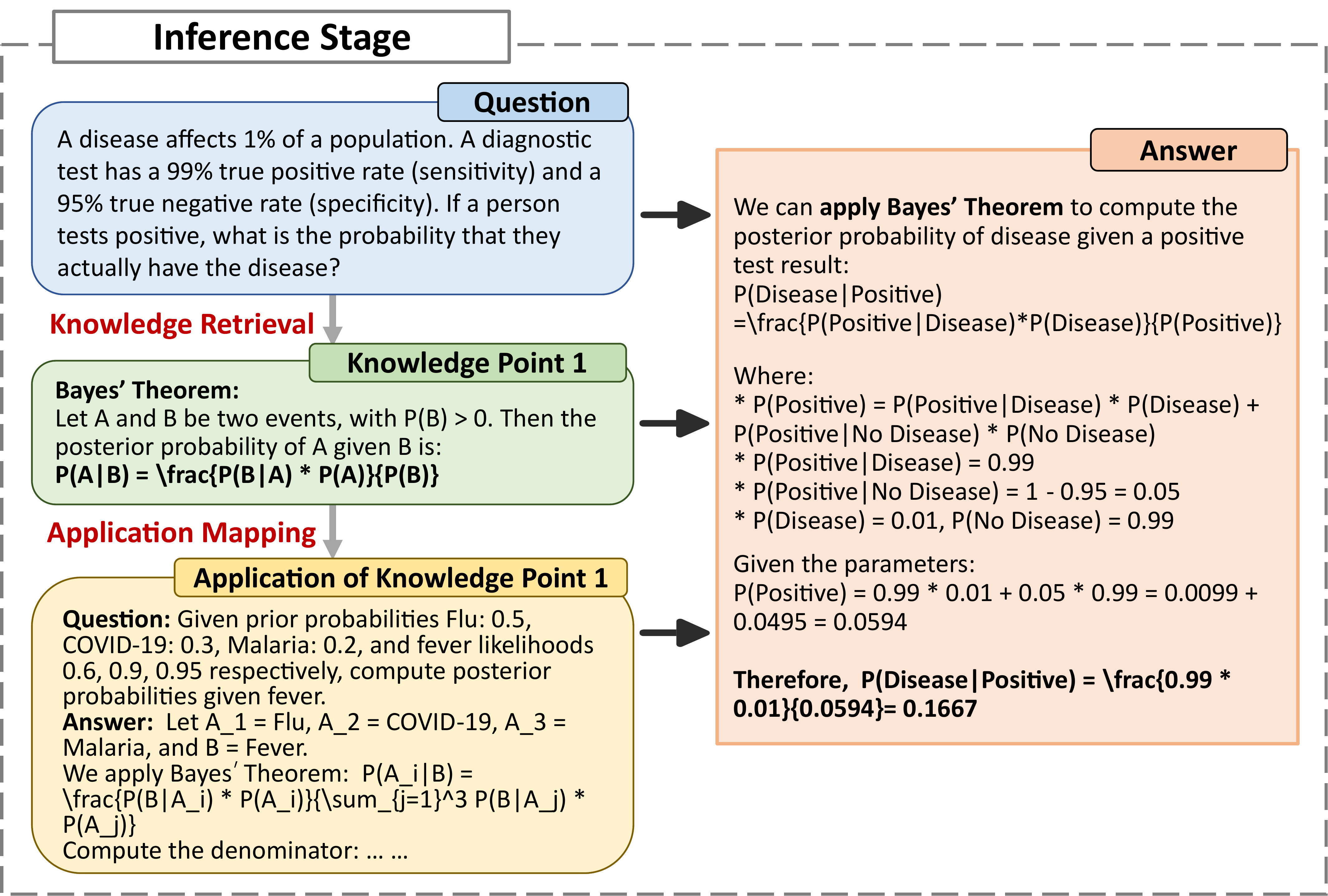}} 
  \caption{The overview framework of RAG+: (a) construction of the application corpus and (b) inference with retrieved knowledge and applications.}

  \label{fig:main}
\end{figure*}

We propose RAG+, a principled extension of Retrieval-Augmented Generation (RAG) that incorporates the explicit application of retrieved knowledge to improve reasoning. While prior RAG frameworks focus on retrieving relevant knowledge for downstream tasks, they often overlook explicitly guiding models on how to utilize this knowledge in reasoning. RAG+ addresses this limitation by introducing an explicit step that applies the retrieved knowledge through aligned application examples, illustrating its practical use.

The RAG+ pipeline consists of two stages: (a) a construction stage, where an application corpus is built and aligned with the knowledge corpus; and (b) an inference stage, in which both knowledge and corresponding applications are retrieved to form a comprehensive prompt for response generation, as shown in Figure~\ref{fig:main}.

\subsection{Construction Stage}

The construction stage aims to build an application corpus $A$ aligned with an existing knowledge corpus $K$, as shown in Figure~\ref{fig:construction}. For each knowledge item $k \in K$, an application example $a \in A$ is either retrieved or generated to demonstrate the practical use of $k$. These examples bridge the gap between passive knowledge access and task-oriented reasoning.

Depending on domain characteristics and data availability, we consider two complementary strategies for constructing application examples: \textit{application generation} and \textit{application matching}.

\textbf{Application Generation:} In many domains, while structured knowledge corpora exist, corresponding application examples remain scarce or incomplete. To address this gap, we leverage powerful LLMs to automatically generate application examples. This process produces a structured application corpus aligned with the knowledge base, facilitating application-aware reasoning. We categorize knowledge items into two types based on their inherent nature to ensure generation of relevant and task-appropriate applications:

\textit{Conceptual knowledge} comprises static, descriptive information, such as definitions, theoretical explanations, or descriptions of entities and principles. The corresponding applications generally involve comprehension tasks, contextual interpretations, or analogies that elucidate meaning and deepen understanding.

\textit{Procedural knowledge} refers to dynamic, actionable information including problem-solving strategies, inference rules, and step-by-step methods. Its associated applications are demonstrated through worked examples, reasoning chains, or practical problem-solving instances where the knowledge is actively applied.

\begin{figure}[t]
  \centering
   \subfigure[An Example of Conceptual Knowledge and Corresponding Applications.]{
    \label{fig:conceptual_app}
    \includegraphics[width=0.495\textwidth]{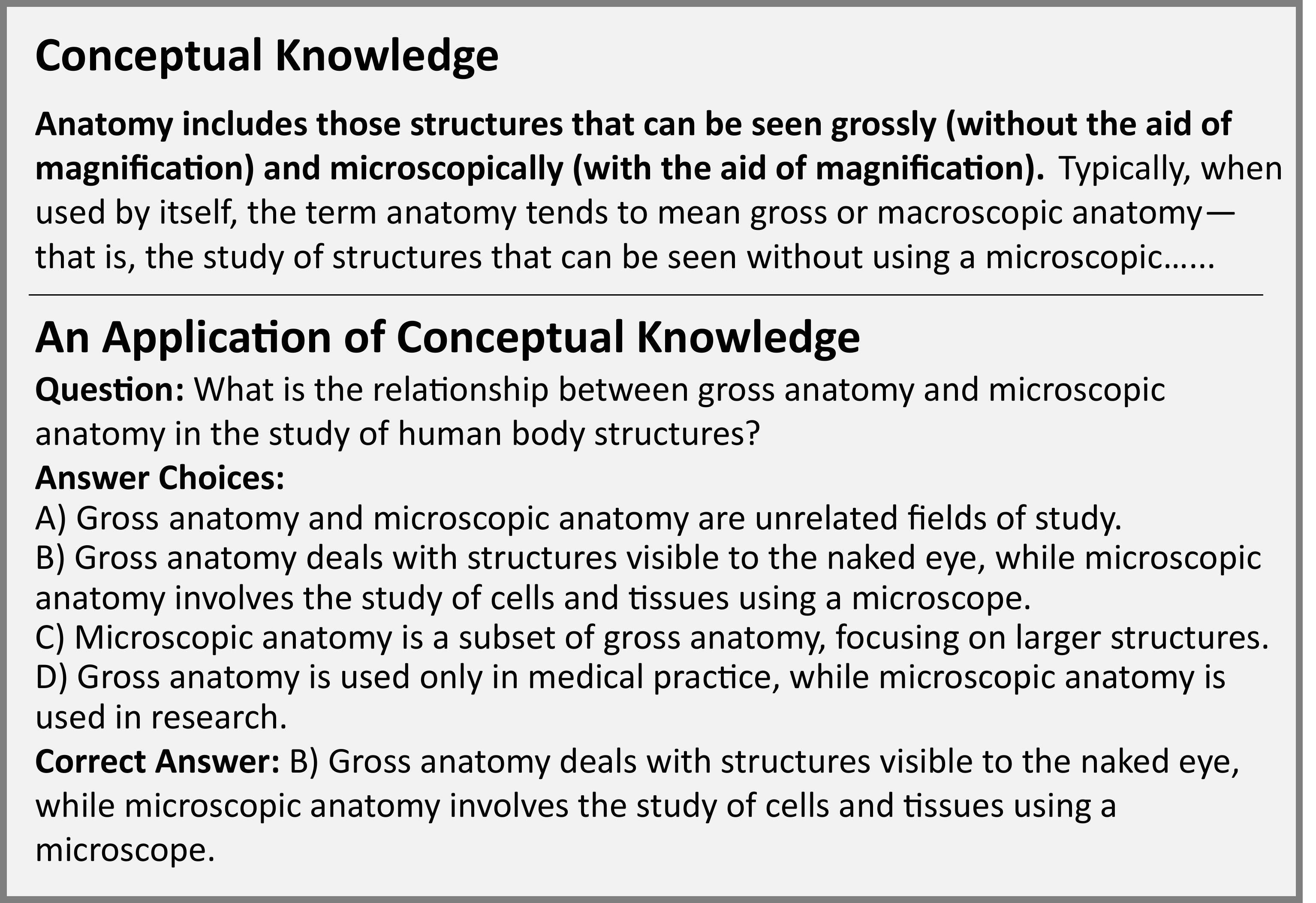}} 
  \subfigure[An Example of Procedural Knowledge and Corresponding Applications.]{
    \label{fig:procedural_app}
    
    \includegraphics[width=0.492\textwidth]{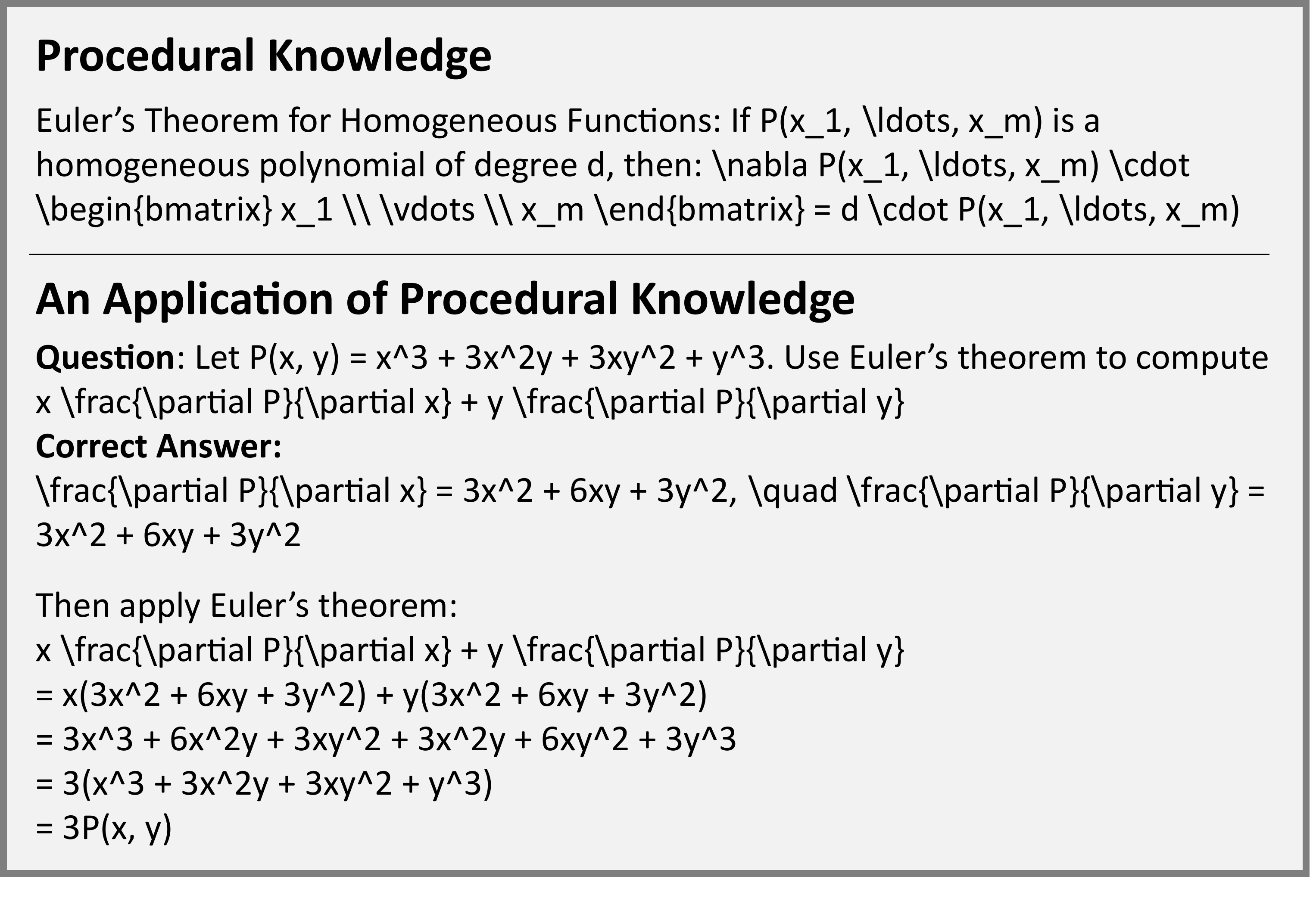}} 
  \caption{Examples of aligning different types of knowledge with application instances.}
  \label{fig:knowledge_applications}
\end{figure}

Figure~\ref{fig:knowledge_applications} illustrates examples of these knowledge types alongside their generated applications. By aligning each knowledge item with a representative application, the constructed corpus enables downstream systems not only to retrieve relevant information but also to engage in more effective, application-aware reasoning.

Guided by the prior classification of knowledge items into conceptual and procedural types, we design tailored prompting strategies to elicit task-appropriate applications: comprehension or contextualization tasks for conceptual knowledge and worked examples or reasoning chains for procedural knowledge.

\textbf{Application Matching:} In domains where real-world cases naturally exemplify the use of specific knowledge, each knowledge item is paired with one or more application instances drawn from authentic scenarios, serving as grounded demonstrations that contextualize and concretize the corresponding knowledge. We refer to these instances as application examples.
To establish these pairings, we first assign both problems and knowledge items to broad semantic categories using powerful LLMs with temperature sampling and self-consistency voting\footnote{The voting mechanism is not essential. To reduce computational cost, a well-designed prompt can serve as an alternative, allowing the alignment of knowledge–application in a single pass.}. Specifically, we first classify knowledge items and application instances separately. Then, within each category, we use LLMs to match each knowledge item with its relevant application examples. The resulting pairings are manually refined to ensure accuracy.
Next, within each category, relevance selection is conducted by prompting the model to identify the most pertinent knowledge entries for each problem. 

This two-stage process yields a many-to-many mapping between knowledge points and application examples, ensuring comprehensive and grounded coverage of relevant applications. For knowledge points lacking matched real-world applications, we supplement them with automatically generated examples as described above, thereby maintaining the completeness and robustness of the application corpus.

These two strategies enable the construction of an application corpus that supports application-aware reasoning across diverse domains.

\subsection{Inference Stage}

During inference, given a test query, RAG+ first retrieves relevant knowledge items from the knowledge corpus based on semantic and structural similarity, using any retrieval method (e.g., dense retrieval, reranking), as shown in Figure~\ref{fig:inference}. For each retrieved knowledge item $k$, its corresponding application example $a$, pre-aligned during the construction stage, is retrieved from the application corpus.
The pair $(k, a)$ is then incorporated into a predefined prompt template that guides the model with both factual information and procedural cues. This prompt is subsequently fed into the language model for final answer generation.

RAG+ is retrieval-agnostic and can be seamlessly integrated into any existing RAG pipeline. Since the knowledge-application alignment is established offline, no modifications to retrieval or generation models are needed at inference. This modular design allows RAG+ to function as a plug-and-play enhancement for diverse reasoning tasks.

\begin{table*}[t]
\centering
\caption{Accuracy (\%) of different models on the MathQA dataset with and without application-level augmentation. ``+'' indicates the use of application-level augmentation.}
\label{tab:na_main_results}
\resizebox{\textwidth}{!}{
\begin{tabular}{l|ccccccc}
\toprule
\textbf{Method} & \textbf{GLM4-9B} & \textbf{Qwen2.5-7B} & \textbf{DS-Qwen-7B} & \textbf{Qwen2.5-14B} & \textbf{DS-Qwen-32B} & \textbf{Qwen2.5-72B} & \textbf{LLaMA3.3-70B} \\ 
\midrule
Baseline & 46.51 & 58.84 & 24.19 & 66.98 & 80.00 & 69.07 & 69.07 \\ 
\midrule
RAG & 47.21 & 64.65 & 24.42 & 73.49 & 82.79 & 70.47 & 71.16 \\
RAG+ & \textbf{52.09} & \textbf{65.58} & \textbf{26.28} & \textbf{74.67} & \textbf{84.65} & \textbf{73.72} & \textbf{71.86} \\ 
\midrule
AFRAG & 48.14 & 63.51 & 22.56 & 66.98 & 82.09 & 71.86 & 70.23 \\
AFRAG+ & \textbf{51.16} & \textbf{64.42} & \textbf{23.95} & \textbf{70.00} & \textbf{83.95} & \textbf{76.05} & \textbf{71.86} \\ 
\midrule
GraphRAG & 33.95 & 56.98 & 27.21 & 69.07 & 82.79 & \textbf{73.02} & 68.37 \\
GraphRAG+ & \textbf{36.51} & \textbf{59.77} & \textbf{33.72} & \textbf{69.77} & \textbf{83.49} & 72.56 & \textbf{69.00} \\ 
\midrule
Rerank RAG & \textbf{48.21} & 56.05 & 26.05 & 71.40 & 80.46 & 73.26 & 74.65 \\
Rerank RAG+ & -- & \textbf{56.28} & \textbf{32.09} & \textbf{78.90} & \textbf{83.26} & \textbf{77.21} & \textbf{76.74} \\
\bottomrule
\end{tabular}
}
\end{table*}

\section{Experiment Setup}
\subsection{Baseline}
To assess the effectiveness of the proposed RAG+ framework, we compare it against several representative RAG-based baselines, each embodying a distinct approach to utilizing retrieved information.
RAG~\cite{lewis2020retrieval} is the standard framework that retrieves relevant documents based on the input query and generates responses conditioned on both the query and the retrieved content. Answer-First RAG (AFRAG) first generates a candidate answer from the query, which is then used to retrieve supporting evidence. The final output is produced using both the original query and the retrieved context. GraphRAG~\cite{edge2024local} incorporates structured knowledge via knowledge graphs to facilitate multi-hop reasoning and improve contextual relevance. Rerank RAG re-ranks the top-$k$ retrieved documents by a large language model and selects the top three for answer generation to enhance query-context alignment.

In the Rerank RAG setup, the same model performs both reranking and generation. Smaller models (e.g., GLM4-9B, DS-Qwen-7B) often fail to comply with reranking instructions and instead generate answers directly. This issue remains despite prompt tuning and leads to missing results on the MathQA and MedQA datasets. Larger models (14B and above), however, reliably perform reranking without this problem.

\subsection{Datasets}

RAG+ is evaluated on three reasoning-intensive domains: mathematics, legal, and medicine.  
The MathQA dataset is constructed from publicly available educational resources and is paired with a custom mathematical knowledge corpus.  
For legal, the sentencing prediction dataset from CAIL 2018~\cite{xiao2018cail2018largescalelegaldataset, zhong2018overviewcail2018legaljudgment} is used, with a knowledge corpus composed of statutes from the Criminal Law of China.  
For medicine, the MedQA dataset~\cite{jin2020disease} is employed, together with a curated medical corpus from \cite{xiong2024benchmarking} relevant to clinical reasoning.

Because the legal and medical corpora lack sufficient real-world applications, we use automatic generation methods for their application corpora. Specifically, generated applications in legal reflect case rulings from the Chinese Criminal Law corpus, while in medicine, knowledge items are categorized before generating aligned applications based on a clinical knowledge base and the MedQA dataset.

In contrast, the math corpus includes authentic application instances, enabling us to employ an application matching approach. This approach employs a two-stage filtering process: first, category alignment assigns both problems and knowledge items to broad categories using Qwen2.5-72B with temperature sampling and self-consistency voting, followed by manual refinement to improve accuracy. Second, relevance selection prompts the model to identify the most pertinent knowledge entries within each category. To maintain corpus completeness, knowledge points without matched real-world applications (under 10\%) are supplemented with automatically generated examples.

The resulting corpus pairs each knowledge item with one or more applications, enabling RAG+ to retrieve both during inference. All prompts used are detailed in Appendix~\ref{app:prompt}.

\subsection{Models}
Nine conversational models from the Qwen~\cite{qwen2025qwen25technicalreport,yang2025qwen3technicalreport}, LLaMA~\cite{dubey2024llama3herdmodels}, DeepSeek~\cite{deepseekai2025deepseekr1incentivizingreasoningcapability}, and ChatGLM~\cite{glm2024chatglm} series were evaluated. Detailed configurations of the evaluated models are provided in Appendix~\ref{app:model}.

\begin{table*}[tp]
\centering
\caption{Accuracy (\%) of different models on sentencing prediction tasks with and without application-level augmentation. LLaMA3.1-8B* denotes the Chinese version, and ``+'' indicates the use of application-level augmentation.}
\label{tab:legal_main_results}
\resizebox{\textwidth}{!}{
\begin{tabular}{l|ccccccc}
\toprule
\textbf{Method} & \textbf{LLaMA3.1-8B*} & \textbf{DS-Qwen-7B} & \textbf{DS-Qwen-32B} & \textbf{QwQ-32B} & \textbf{Qwen3-32B} & \textbf{Qwen2.5-72B} & \textbf{LLaMA3.3-70B} \\ \midrule
Baseline & 29.00 & 53.00 & 80.50 & 80.00 & 73.00 & 73.00 & 51.50 \\ \midrule

RAG & 36.00 & 65.50 & \textbf{85.50} & 81.50 & \textbf{83.00} & 76.50 & 70.50 \\
RAG+ & \textbf{41.00} & \textbf{67.50} & \textbf{85.50} & \textbf{86.00} & 82.50 & \textbf{83.00} & \textbf{76.00} \\ \midrule

AFRAG & 27.50 & 65.50 & 85.00 & 82.50 & 76.00 & 85.00 & 41.50 \\
AFRAG+ & \textbf{33.00} & \textbf{68.00} & \textbf{86.50} & \textbf{83.00} & \textbf{77.50} & \textbf{86.50} & \textbf{52.50} \\ \midrule

GraphRAG & 36.50 & 42.00 & \textbf{81.50} & 76.00 & 68.50 & \textbf{64.00} & 38.50 \\
GraphRAG+ & \textbf{46.00} & \textbf{47.50} & \textbf{81.50} & \textbf{77.50} & \textbf{75.00} & \textbf{64.00} & \textbf{52.00} \\ \midrule

Rerank RAG & 33.00 & 60.00 & 82.00 & \textbf{83.50} & 80.50 & 77.50 & \textbf{77.50} \\
Rerank RAG+ & \textbf{34.00} & \textbf{61.00} & \textbf{82.50} & \textbf{83.50} & \textbf{82.00} & \textbf{87.50} & \textbf{77.50} \\
\bottomrule
\end{tabular}
}
\end{table*}

\section{Results}
In this section, the effectiveness of RAG+ is demonstrated across multiple models on three reasoning-intensive datasets spanning distinct domains. Performance trends are also analyzed across different model scales within the Qwen2.5 series. Ablation studies examine the effects of incorporating only application examples or using larger models for reranking. Finally, case studies illustrate how RAG+ enhances complex reasoning.

\begin{table*}[tp]
\centering
\caption{Accuracy (\%) of different models on the MedQA dataset with and without application-level augmentation. ``+'' indicates the use of application-level augmentation.}
\label{tab:med_main_result}
\resizebox{\textwidth}{!}{
\begin{tabular}{l|ccccccc}
\toprule
\textbf{Method} & \textbf{LLaMA3.1-8B} & \textbf{Qwen2.5-7B} & \textbf{DS-Qwen-7B} & \textbf{DS-Qwen-32B} & \textbf{QwQ-32B} & \textbf{Qwen2.5-72B} & \textbf{LLaMA3.3-70B} \\ \midrule
Baseline & 57.80 & 41.80 & 32.60 & 80.00 & 80.40 & 73.80 & 78.20 \\ \midrule

RAG & 63.00 & \textbf{59.20} & 34.60 & 79.00  & 80.20 & 75.00 & 80.20 \\
RAG+ & \textbf{63.60} & 57.60 & \textbf{40.20} & \textbf{80.20} & \textbf{80.80} & \textbf{75.40} & \textbf{81.40} \\ \midrule

AFRAG & 56.40 & 53.40 & 32.20 & 78.20 & 81.20 & 76.40 & 82.40 \\
AFRAG+ & \textbf{57.00} & \textbf{57.20} & \textbf{34.60} & \textbf{78.60} & \textbf{82.20} & \textbf{77.40} & \textbf{83.00} \\ \midrule

Rerank RAG & 60.00 & 58.60 & \textbf{35.20} & 79.80 & 80.60 & 76.40 & 81.00 \\
Rerank RAG+ & \textbf{63.40} & \textbf{61.40} & -- & \textbf{80.20} & \textbf{81.40} & \textbf{78.20} & \textbf{85.60} \\
\bottomrule
\end{tabular}
}
\end{table*}

\subsection{Mathematics Domain}

Table~\ref{tab:na_main_results} reports the model performance on the MathQA dataset with and without application-level augmentation. Most retrieval methods demonstrate effectiveness, though some negatively impact performance. Nevertheless, almost all augmented variants outperform their non-augmented counterparts.

Notably, Qwen2.5-14B achieves a substantial improvement of over 7.5\% with Rerank RAG+, while DS-Qwen-7B showing gains of 6.5\% and 6.0\% with GraphRAG+ and Rerank RAG+, respectively. GLM4-9B and Qwen2.5-72B show consistent gains between 2.8\% and 4.8\% across multiple RAG+ variants. In contrast, AFRAG and GraphRAG tend to be less effective on smaller models, such as Qwen2.5-14B with GraphRAG, likely because these methods depend heavily on a model’s ability to interpret complex relational structures and integrate them into reasoning, which smaller models often lack.

Additionally, GraphRAG without application augmentation can lead to performance drops. For instance, Qwen2.5-72B shows a slight decline with plain GraphRAG, possibly due to its emphasis on entity definitions and relations may not align well with mathematical tasks that require solution-oriented knowledge such as formulas.

Overall, most models achieve accuracy improvements between 2.5\% and 6.5\%, while the overall range spans from approximately 0.7\% to 7.5\%. These findings highlight the value of application-aware augmentation in reasoning tasks that require more than factual knowledge.

\subsection{Legal and Medicine Domain}

Table~\ref{tab:legal_main_results} and Table~\ref{tab:med_main_result} show the performance of various RAG-based methods on the sentencing prediction task and the MedQA dataset across different models. Application-level augmentation consistently improves accuracy over both base models and standard RAG variants.

In the legal domain, Qwen2.5-72B achieves 87.5\% accuracy with Rerank RAG+, a 10\% gain over its non-augmented version. DS-Qwen-32B and QwQ-32B also show notable improvements with RAG+ and AFRAG+, demonstrating the effectiveness of application-aware augmentation. In contrast, GraphRAG alone underperforms, especially for smaller models like DS-Qwen-7B and LLaMA3.1-8B, likely due to its focus on entity-level information. However, combining GraphRAG with application augmentation significantly improves results, highlighting the need for task-aligned retrieval. LLaMA3.3-70B performs well across all methods but shows marginal gains with application augmentation, indicating diminishing returns from retrieval augmentation as model size increases.

On the MedQA dataset, Rerank RAG+ yields the best performance for most models, especially the larger ones. For example, LLaMA3.3-70B reaches 85.6\%, surpassing its baseline (81\%) and Rerank RAG (81.0\%) methods. Smaller models like Qwen2.5-7B and LLaMA3.1-8B also benefit, with gains of 2.2\% and 3.4\%, respectively. AFRAG and its augmented version provide steady improvements, showing that grounding abstract medical knowledge through applications enhances reasoning. While standard RAG offers a solid boost, application-level augmentation is essential for maximizing performance across model sizes.

These results demonstrate the consistent effectiveness of application-aware augmentation in enhancing RAG-based methods, benefiting both large and small models when paired with effective retrieval strategies.

\subsection{Effect of Model Scale}

\begin{figure}[t]
\centering
\includegraphics[width=0.499\textwidth]{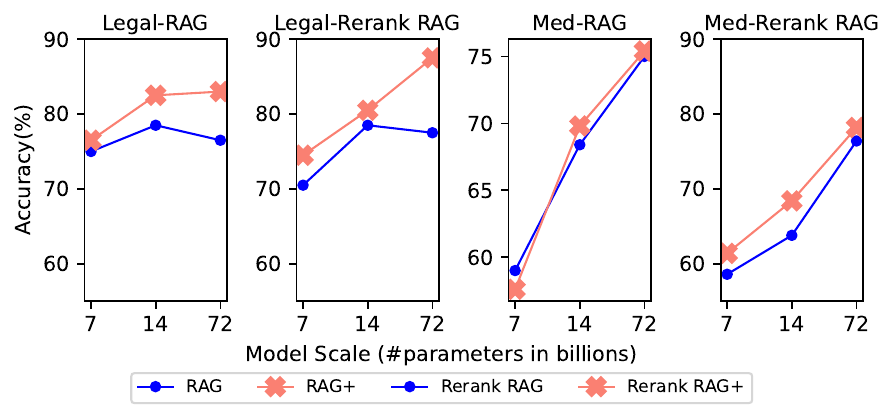}
\caption{Performance comparison on sentencing prediction tasks and the MedQA dataset using Qwen2.5 series models across different scales with and without application examples.}
\label{fig:modelscale}
\end{figure}

As shown in Figure~\ref{fig:modelscale}, all methods show consistent performance gains as model size increases from 7B to 14B and then to 72B, reflecting the enhanced reasoning capabilities of larger models. Notably, approaches augmented with application-level examples achieve larger improvements than their non-augmented counterparts. 

In the legal domain, accuracy rises steadily with model scale when application augmentation is used. In the medicine domain, while all models benefit, the performance gains from scaling are less pronounced. These results indicate that application-guided retrieval scales especially well with model size, further boosting knowledge integration and reasoning in complex, domain-specific tasks.

\subsection{Impact of Reranking Model}

\begin{table}[tp]
\centering
\small
\caption{Accuracy (\%) on sentencing prediction tasks across different RAG methods. ``Rerank (72B)'' denotes reranked by Qwen2.5-72B, generated by base model.}
\label{tab:sentencing_72b}
\begin{tabular}{l|cc}
\toprule
\textbf{Method} & \textbf{Qwen2.5-7B} & \textbf{Qwen2.5-14B} \\ \midrule
Baseline & 46.00 & 74.00 \\ \midrule
RAG & 75.00 & 78.50 \\
RAG+ & \textbf{76.50} & \textbf{82.50} \\ \midrule
Rerank RAG & 70.50 & 78.50 \\
Rerank RAG+ & \textbf{74.50} & \textbf{81.00} \\ \midrule
Rerank (72B) RAG & 72.50 & 80.00 \\
Rerank (72B) RAG+ & \textbf{83.50} & \textbf{86.00} \\ \bottomrule
\end{tabular}
\end{table}

\begin{table}[tp]
\small
\centering
\caption{Accuracy (\%) on the MedQA dataset across different RAG methods.}
\label{tab:medqa_72b}
\begin{tabular}{l|cc}
\toprule
\textbf{Method} & \textbf{Qwen2.5-7B} & \textbf{Qwen2.5-14B} \\ \midrule
Baseline & 41.80 & 68.80 \\ \midrule
RAG & \textbf{59.20} & 68.40 \\
RAG+ & 57.60 & \textbf{69.80} \\ \midrule
Rerank RAG & 58.60 & 63.80 \\
Rerank RAG+ & \textbf{61.40} & \textbf{68.40} \\ \midrule
Rerank (72B) RAG & 60.80 & 69.80 \\
Rerank (72B) RAG+ & \textbf{61.80} & \textbf{70.00} \\ \bottomrule
\end{tabular}
\end{table}

To investigate the suboptimal performance of Rerank RAG on smaller models, we conducted an ablation study by replacing the initial reranker with a stronger model, Qwen2.5-72B, to assess its impact on downstream accuracy.

As shown in Table~\ref{tab:sentencing_72b} and Table~\ref{tab:medqa_72b}, we isolate the effect of reranking quality by varying the reranker while keeping other components fixed. Stronger reranking consistently improves accuracy across both tasks, with the largest gains in the Rerank (72B) RAG+ setting, where application-augmented reranking achieves the best performance. Notably, its improvement over non-augmented reranking exceeds that between Rerank RAG+ and Rerank RAG, highlighting the joint benefits of reranking strength and application-level guidance. These results underscore the value of cross-model collaboration, where large models enhance retrieval to support more efficient inference in smaller ones.

\begin{figure*}[t]
\centering
\includegraphics[width=\textwidth]{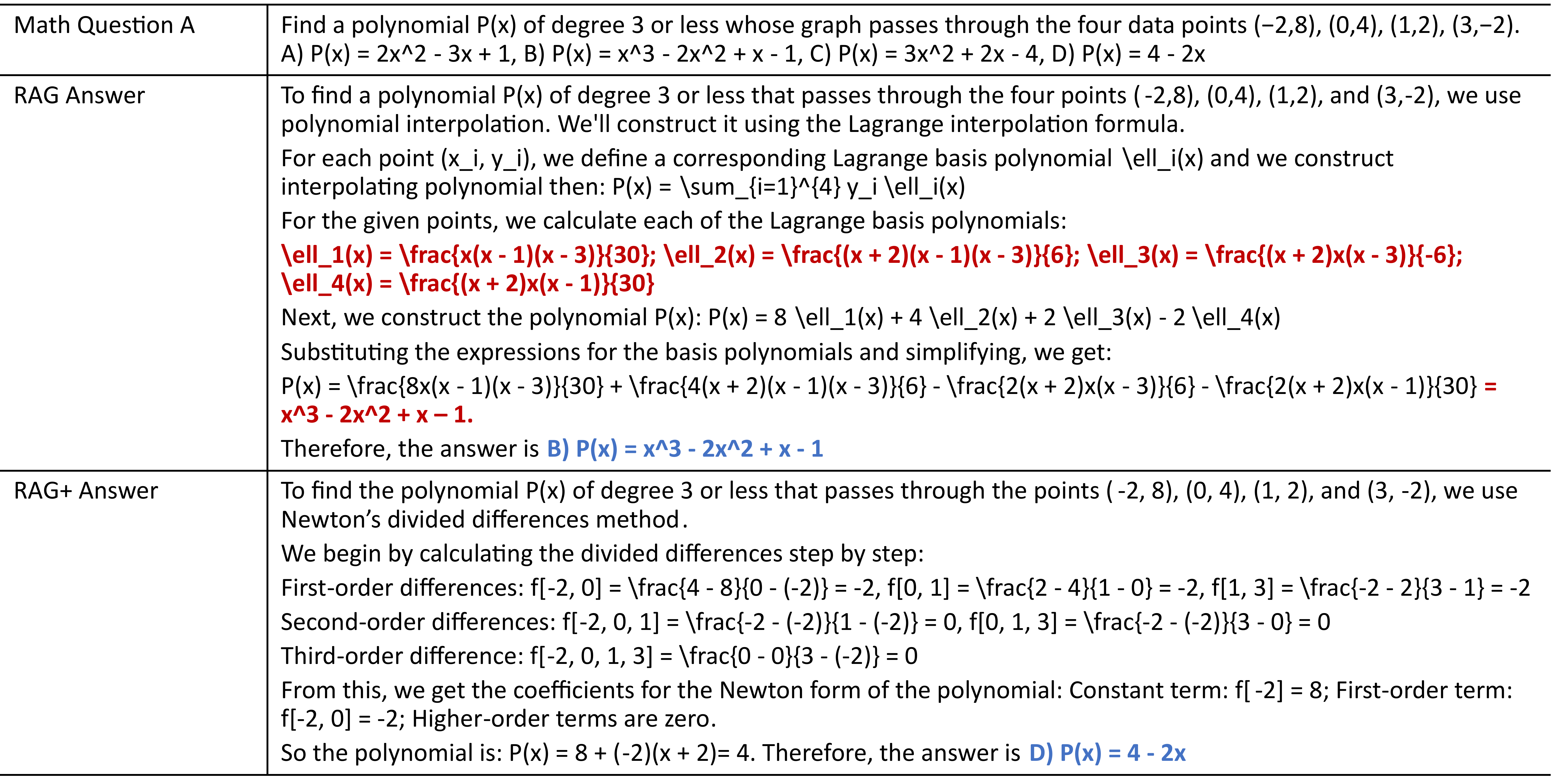}
\caption{Comparison of output reasoning paths between RAG and RAG+ in the math domain using the Qwen2.5-72B model.}
\label{fig:casestudymath}
\end{figure*}

\subsection{Effect of Application Examples Only}

We conducted experiments to assess the contribution of application examples without their associated knowledge on the legal and math domains. Results are shown in Table~\ref{tab:legal-rag} and Table~\ref{tab:math-rag}. Incorporating only application examples leads to performance improvements over the base models in most cases. This indicates that application-level guidance alone provides a meaningful benefit. However, this setting still underperforms compared to the full configuration where both knowledge and application examples are included, referred to as RAG+ and AFRAG+.

These findings suggest that while application examples alone can partially support reasoning, likely by providing structural cues or task-specific formulations, the explicit inclusion of knowledge remains essential for achieving optimal performance. Therefore, combining knowledge with its applications produces the most effective results.

\begin{table}[tp]
\centering
\caption{Accuracy (\%) on legal tasks across different RAG methods. ``--'' indicates application-only retrieval; ``+'' indicates the inclusion of retrieved knowledge and application.}
\label{tab:legal-rag}
\resizebox{0.5\textwidth}{!}{
\begin{tabular}{l|ccc}
\toprule
\textbf{Method} & \textbf{Qwen2.5-7B} & \textbf{Qwen2.5-14B} & \textbf{DS-Qwen-32B}  \\ \midrule
Baseline & 46.00 & 74.00 & 80.50  \\ \midrule
RAG- & 65.00 & 76.50 & 84.50  \\
RAG & 75.00 & 78.50 & \textbf{85.50}  \\
RAG+ & \textbf{76.50} & \textbf{82.50} & \textbf{85.50}  \\ \midrule

AFRAG- & 34.50 & 74.50 & 83.00 \\
AFRAG & 40.50 & 81.00 & 85.00  \\
AFRAG+ & \textbf{47.00} & \textbf{85.50} & \textbf{86.50}  \\ \bottomrule
\end{tabular}
}
\end{table}

\begin{table}[tp]
\centering
\caption{Accuracy (\%) on math tasks across different RAG methods. ``--'' indicates application-only retrieval; ``+'' indicates the inclusion of retrieved knowledge and application.}
\label{tab:math-rag}
\resizebox{0.5\textwidth}{!}{
\begin{tabular}{l|ccc}
\toprule
\textbf{Method} & \textbf{Qwen2.5-7B} & \textbf{Qwen2.5-14B} & \textbf{Qwen2.5-72B}  \\ \midrule
Baseline & 58.84 & 66.98 & 69.07  \\ \midrule
RAG- & 58.19 & 69.53 & 69.07  \\
RAG & 64.65 & 73.49 & 70.47  \\
RAG+ & \textbf{65.58} & \textbf{74.67} & \textbf{73.72}  \\ \midrule

AFRAG- & 59.81 & 66.98 & 74.19 \\
AFRAG & 63.51 & 66.98 & 71.86  \\
AFRAG+ & \textbf{64.42} & \textbf{70.00} & \textbf{76.05}  \\ \bottomrule
\end{tabular}
}
\end{table}

\subsection{Effect of Open-Ended Experiments}
\begin{table}[tp]
\small
\centering
\caption{Accuracy (\%) of open-ended sentencing prediction results.}
\label{tab:open-ended}
\resizebox{0.48\textwidth}{!}{
\begin{tabular}{l|ccc}
\toprule
\textbf{Method} & \textbf{Qwen2.5-7B} & \textbf{Qwen2.5-14B} & \textbf{Qwen2.5-72B}  \\ \midrule
Baseline & 49.00 & 75.50 & 85.00  \\ \midrule
RAG & 	61.50 & 75.50 & 89.50  \\
RAG+ & \textbf{66.50} & \textbf{76.50} & \textbf{90.50}  \\ \midrule

AFRAG & 63.00 & 71.00 & 83.50  \\
AFRAG+ & \textbf{66.50} & \textbf{79.50} & \textbf{86.00}  \\ \bottomrule
\end{tabular}
}
\end{table}

We additionally conducted open-ended experiments on the legal sentencing prediction dataset. In this setting, the model is asked to generate a free-form sentence length prediction (e.g., “6–12 months”), and the output is considered correct if it falls within the gold interval. The results are summarized in Table~\ref{tab:open-ended}. RAG+ consistently improves performance across all model sizes, demonstrating robustness even under open-ended evaluation. This further supports the central hypothesis that combining knowledge with aligned application examples enhances reasoning and task performance.

\subsection{Case Study}
To evaluate the practical effectiveness of RAG+, we present a case study on the mathematics domain, as illustrated in Figure~\ref{fig:casestudymath}. The answer generated by RAG correctly identifies the interpolation method as Lagrange interpolation but fails to execute it accurately due to the complexity of intermediate symbolic expressions. While the approach is mathematically valid, errors in deriving the basis polynomials lead to an incorrect final result. 
In comparison, the Newton divided differences method, although less commonly emphasized in retrieval-based settings, provides a more transparent and step-by-step procedure. Its recursive computation of coefficients reduces algebraic errors and produces the correct polynomial. This suggests that even when the correct method is retrieved, symbolic reasoning may fail due to execution errors, highlighting the need for verification mechanisms alongside retrieval.

\section{Discussion}

To enable application-aware retrieval, we propose a lightweight method for constructing application-specific corpora from a knowledge base, typically generating one to two examples per item. The corpus grows linearly with the number of items and introduces no retrieval overhead, as each item is directly paired with an application example.
In experiments, the initial corpus sizes were 223 KB (math), 528 KB (legal), and 99,382 KB (medicine). Generating the medicine corpus with Qwen2.5-72B on eight 64 GB NPUs took around six hours, which is acceptable given this scale. Final sizes reached 612 KB, 868 KB, and 105,558 KB, respectively, closely matching the sizes of the underlying knowledge corpora. The method is scalable and supports efficient incremental updates.

\section{Conclusion}

In this work, we introduce RAG+, a framework that integrates application-level augmentation into retrieval-augmented generation. Through comprehensive experiments across diverse domains and model scales, we demonstrate that incorporating application examples consistently leads to performance improvements.
RAG+ outperforms baselines across tasks and RAG variants, highlighting the value of structured application in leveraging retrieved knowledge for reasoning.
Our results indicate that retrieval alone is insufficient—effective alignment and application of retrieved knowledge are crucial. Future work may explore more advanced application strategies and tighter integration to further enhance reasoning in LLMs.

\section*{Limitations}

While RAG+ consistently improves performance, it also has several limitations. First, constructing a high-quality application corpus can be resource-intensive, especially in domains with limited annotated data. Automated generation depends heavily on large language models, which may introduce errors or oversimplify complex reasoning.

Second, RAG+ assumes a strong alignment between knowledge and application pairs, but mismatches can occur—particularly when retrieved knowledge is noisy or incomplete—leading to incorrect or misleading reasoning.

Finally, our current approach focuses on enhancing reasoning via application-level augmentation, but does not directly address retrieval quality or efficiency, which remain critical to overall performance. Future work should explore joint optimization of retrieval and application generation, as well as better handling of uncertainty and ambiguity in retrieved content.

\section*{Ethics Statement}
All datasets and models used in this study are open-source, and the licenses of the models have been clearly specified. The prompts used in the experiments, along with the full experimental environment and all relevant parameters, are provided to ensure reproducibility. 
The entire study can be replicated using widely available large model API frameworks. Furthermore, all manual operations involved in the process were carried out solely by the authors, without the involvement of any external collaborators or paid contributors. This guarantees both the transparency and independence of the research.

\section*{Acknowledgement}
This work was supported by National Natural Science Foundation of China (62232014, 62272377).

\bibliography{anthology,custom}
\bibliographystyle{acl_natbib}

\clearpage
\appendix

\section{Experimental Details}
In this study, we employ a range of hardware devices, including units of 32 GB Ascend 910B4, 64 GB Ascend 910B3, and 32 GB Tesla V100 PCIe units. 
For large-scale models exceeding 70 billion parameters (e.g., LLaMA3.3-70B and Qwen2.5-72B), we deploy eight Ascend 910B4 devices (32 GB each). For 32B-scale models (e.g., Qwen2.5-32B, Qwen3-32B, QwQ-32B, and DS-Qwen-32B), two Ascend 910B4 devices are deployed. 
Models with approximately 14 billion parameters (e.g., Qwen2.5-14B) are run on a single 64 GB Ascend 910B3, while models around the 8B scale (e.g., Llama3.1-8B, Qwen2.5-7B, DS-Qwen-7B, and ChatGLM4-9B) are executed on a single 32 GB Ascend 910B4. 
For models used in retrieval-augmented generation (RAG) components (e.g., bg3-m3 and bge-reranker-v2-m3), we utilize a 32 GB Tesla V100 PCIe. All deployments are carried out using the vLLM inference framework~\cite{kwon2023efficient}.

For models generating long-form outputs (i.e., QwQ-32B, DS-Qwen-7B, and DS-Qwen-32B), we use stochastic decoding (temperature = 1, top\_p = 1) to promote output diversity. For all other models, we apply a deterministic decoding strategy with a temperature of 0 and top\_p of 1. Each experiment is conducted three times, and we report the average performance across these runs.

To implement RAG functionality, we utilize the Dify framework. Specifically, Dify is used for corpus uploading, with a segmentation length of 600 tokens. The text is pre-segmented to ensure that the divisions are structurally coherent and suitable for retrieval.

\subsection{Model sources and licenses}\label{app:model}
This study employs nine publicly available conversational models: LLaMA-Chinese-8B-Instruct (sourced from ModelScope) and LLaMA3.1-8B, LLaMA3.3-70B, Qwen2.5-7B, Qwen2.5-72B, QwQ-32B, DeepSeek-R1-Distill-Qwen-7B, DeepSeek-R1-Distill-Qwen-32B, and ChatGLM4-9B (all obtained from Hugging Face). Additionally, the BGE-M3 embedding model~\cite{bge-m3} is utilized for generating text embeddings. Details regarding model sources and licensing are provided in Table~\ref{tab:modelsources}.

The LLaMA-Chinese-8B-Instruct model is exclusively used for the sentencing prediction task in the legal domain, as this task is conducted in Chinese. For the remaining two tasks, LLaMA3.1-8B is employed. The QwQ-32B model is excluded from the numerical analysis task due to the high complexity of the questions, which frequently lead to excessively long outputs and potential infinite loops. These issues significantly increase inference time and GPU memory usage, thereby exceeding available computational resources.

The BGE-M3 model is employed to generate embeddings for corpus texts, while the BGE-reranker-v2-m3 model is employed to reorder candidate documents based on their relevance to the user query. This reranking step enhances the overall quality of the semantic retrieval process.

\begin{table*}[htp]
\centering
\small
\renewcommand{\arraystretch}{1.6}
\caption{Models, sources and licenses used in this work.} \label{tab:modelsources}
\begin{tabularx}{0.95\textwidth}{l|l|c}
\toprule
\textbf{Model} & \textbf{URL} & \textbf{Licenses} \\ \midrule 
LLaMA3-Chinese-8B & \renewcommand{\arraystretch}{1.05}\begin{tabular}[c]{@{}l@{}}https://www.modelscope.cn/models/FlagAlpha/\\ LLaMA3-Chinese-8B-Instruct/summary\end{tabular} & Apache License 2.0 \\
LLaMA3.1-8B & https://huggingface.co/meta-llama/Meta-Llama-3.1-8B-Instruct & llama3.1 license \\
LLaMA3.3-70B & https://huggingface.co/meta-llama/Llama-3.3-70B-Instruct & llama3.3 license \\
Qwen2.5-7B & https://huggingface.co/Qwen/Qwen2.5-7B-Instruct & Apache License 2.0 \\
Qwen2.5-72B & https://huggingface.co/Qwen/Qwen2.5-72B-Instruct & Qwen license \\
QWQ-32B & https://huggingface.co/Qwen/QwQ-32B & Apache License 2.0 \\
DS-Qwen-7B & https://huggingface.co/deepseek-ai/DeepSeek-R1-Distill-Qwen-7B & MIT License \\
DS-Qwen-32B & https://huggingface.co/deepseek-ai/DeepSeek-R1-Distill-Qwen-32B & MIT License \\
ChatGLM4-9b & https://huggingface.co/THUDM/glm-4-9b-chat & glm-4-9b License \\
bg3-m3 & https://huggingface.co/BAAI/bge-m3 &  MIT License\\
bge-reranker-v2-m3 & https://huggingface.co/BAAI/bge-reranker-v2-m3 & Apache License 2.0 \\
\bottomrule
\end{tabularx}
\end{table*}

\subsection{Baseline Details}\label{app:baseline}
We evaluate nine configurations: Baseline, RAG~\cite{lewis2020retrieval}, RAG+, Answer-First RAG, Answer-First RAG+, GraphRAG~\cite{edge2024local}, GraphRAG+, Rerank RAG, and Rerank RAG+.

The RAG configuration follows the original implementation by \citeauthor{lewis2020retrieval}~\cite{lewis2020retrieval}, and is implemented using the Dify framework. We employ the BGE-M3 model to embed both corpus chunks and key terms for similarity matching, and the BGE-Reranker-v2-M3 model to rerank the retrieved candidates. High-quality indices are constructed for retrieval, with each corpus chunk limited to 800 tokens. For all RAG-based configurations, the top three most relevant text chunks are retrieved for each query.

\textbf{Baseline}: The Baseline configuration directly uses the large language model (LLM) to answer questions without incorporating any external knowledge retrieval. It serves as the baseline for evaluating the effectiveness of various RAG-based enhancements.

\textbf{RAG}: Vanilla RAG directly applies the retrieval-augmented generation method. The matching fields are the question and its answer options. Retrieved content is incorporated into a prompt template and passed to the model to generate an answer.

\textbf{Answer-First RAG}: Following the approach by \citeauthor{asai2023self}~\cite{asai2023self}, we adopt a more direct implementation. Given a question $Q$, the model $M$ is first prompted to generate a preliminary answer $A^*$, which is then used as a query to retrieve relevant content $C$. The final answer $A$ is generated by prompting the model with both $Q$ and $C$.

\textbf{GraphRAG}: Based on \citeauthor{edge2024local}~\cite{edge2024local}, GraphRAG integrates structured knowledge into a prompt for answer generation. We use Qwen2.5-72B for entity extraction, relationship extraction, claim identification, summary generation, and community report creation. The embedding model is set to ``cl100k\_base'', while ``nomic/text-embedding-nomic-embed-text-v1.5@q4\_k\_m'' is used for graph construction. We adopt a chunk size of 1200 tokens and a maximum cluster size of 10, and use ``local\_search'' for graph retrieval. Only the top relevant chunk is returned during inference.
Due to the large size of the MedQA corpus in the medical domain, we omit GraphRAG for this task, as graph construction and retrieval are prohibitively expensive. However, for the NA and legal domains (each corpus approximately 1000 KB), GraphRAG is feasible and conducted accordingly.

\textbf{Rerank RAG}: Rerank RAG is a multi-step retrieval process. First, vanilla RAG retrieves the top ten candidates. These are then passed, along with the question, to the target model for re-ranking. The top three items from this reranked list are used to construct the final prompt for answer generation. This process allows the model to participate in the retrieval pipeline, enabling more informed selection of relevant knowledge.

\textbf{RAG+} and \textbf{Answer-First RAG+}: In these variants, each retrieved knowledge item is mapped to a corresponding application. Both the knowledge and its application are inserted into the prompt template to guide the model’s answer generation.

\textbf{GraphRAG+}: GraphRAG+ extends GraphRAG by enriching retrieved knowledge with application content. Since constructing dedicated applications for GraphRAG-extracted entities and relations would be labor-intensive, we instead reuse applications from the original corpus via fuzzy matching. Once a match is found, both the application and the matched knowledge—along with the original GraphRAG output—are included in the prompt.

\textbf{Rerank RAG+}: Rerank RAG+ extends Rerank RAG by requiring the model to output results in a specific format, enabling application mapping. However, models frequently fail to follow this format, often providing direct answers or returning too few knowledge items. This inconsistency complicates the parsing process. To address this, we increase the number of runs and re-prompt the model to ensure successful output parsing.
This approach proves effective across most models in our experiments, with the exception of GLM4-9B on the Numerical Analysis dataset.

\subsection{Dataset Details}\label{app:datasets}
\subsubsection{Math QA in Mathematic Domain}
We conduct experiments in the domain of Numerical Analysis. Most existing mathematics datasets are relatively simple—often achieving over 90\% accuracy with current models—and primarily emphasize reasoning rather than Retrieval-Augmented Generation (RAG). To better suit our research objectives, we constructed a new Numerical Analysis dataset from scratch.

We first collected a range of publicly available online resources, supplemented with original questions and corresponding answers. The dataset includes both standard QA items and multiple-choice questions with varying numbers of answer options. Only the correct answers were retained. For compound questions that inquire about multiple values (e.g., "What is the value of $y$ when $x = 0$ or $x = 1$?"), we decomposed them into separate single-answer questions, such as "What is the value of $y$ when $x = 0$" and "What is the value of $y$ when $x = 1$?".

Next, we transformed all QA pairs into a multiple-choice format by prompting powerful LLMs to generate three additional plausible but incorrect answer options. This step expands the test set while simultaneously constraining the model’s generative space.

Finally, we manually reviewed and refined the knowledge points, solution demonstrations, and generated answer choices to ensure overall quality and accuracy. All annotations and validations were conducted solely by the authors without external assistance.

The final dataset consists of 430 test questions in the domain of Numerical Analysis.

\subsubsection{Sentencing Prediction in Legal Domain}
We use data from the CAIL 2018 dataset~\cite{xiao2018cail2018largescalelegaldataset, zhong2018overviewcail2018legaljudgment}, selecting 200 samples from our training set. We focus exclusively on questions related to Article 234 of the Criminal Law, which concerns sentencing for intentional injury. All questions were converted to a multiple-choice format, with answer options corresponding to sentencing durations: less than three years, three to ten years, and more than ten years.

\subsubsection{MedQA in Medicine Domain}

We use the MedQA dataset curated by \citeauthor{jin2020disease}. We randomly sampled 500 examples from the dataset to serve as our training set. All selected items are in a multiple-choice format.

\subsection{Corpus Details}

\subsubsection{Mathematics}

The knowledge points in the Numerical Analysis (NA) corpus are collected from various online sources. These include definitions, theorems, lemmas, factual statements, and methods, and are typically concise. Due to inconsistent formatting, we used Qwen2.5-72B to extract key knowledge points and normalize them into a unified format, as illustrated in Figure~\ref{appfig:math-knowledge-generation}.

After processing, we obtained a total of 816 knowledge points. We examined their lengths with a predefined chunk size of 800 tokens. No knowledge point exceeded this threshold, so no additional splitting was required.

We followed the Dify framework to construct the knowledge base: processed knowledge points were uploaded using the pre-segmented chunks, embedded using the BGE-M3 model, and indexed for high-quality retrieval. During inference, the top three most relevant chunks were retrieved.

\subsubsection{Legal}

The legal corpus is extracted from the Criminal Law of the People's Republic of China. Chunking was based on the natural structure of the legal text, treating each article as a single knowledge point, resulting in 451 items. The corpus is divided into two major sections: the General Provisions (101 items), which provide conceptual knowledge such as definitions and factual descriptions, and the Specific Provisions, which provide solution knowledge in the form of sentencing guidelines.

No further processing was applied. We used a chunk size of 800 tokens and followed the same Dify-based uploading, embedding, and retrieval pipeline as in the mathematics domain.

\subsubsection{Medicine}

The medical corpus is sourced from \citeauthor{xiong2024benchmarking}~\cite{xiong2024benchmarking}. We use only the textbook portion of the corpus, where knowledge is already structured into discrete knowledge points, each representing a self-contained fragment.

The corpus spans 18 subjects and includes 64,117 knowledge points, with a total size of 99,382 KB. Each knowledge point, averaging less than 600 words, was treated as a single chunk without further segmentation. The knowledge encompasses both conceptual and procedural content.

We used a chunk size of 800 tokens, consistent with other domains, and uploaded the data to the Dify system using the same embedding and retrieval procedure.

\section{Experimental Results}
\subsection{Complete Results}
Accuracy is used as the evaluation metric. We conducted three independent inference runs for all experiments and calculated the average results. The complete results are shown in Table~\ref{tab:legal_main_results_full} -~\ref{tab:na_main_results_full}.
\begin{table*}[tp]
\centering
\small
\caption{Accuracy (\%) of different models on the sentencing prediction task with and without application-level augmentation. LLaMA3.1-8B* denotes the Chinese version.}
\label{tab:legal_main_results_full}
\resizebox{\textwidth}{!}{
\begin{tabular}{l|ccccccc}
\toprule
\textbf{Method} & \textbf{LLaMA3.1-8B*} & \textbf{DS-Qwen-7B} & \textbf{DS-Qwen-32B} & \textbf{QwQ-32B} & \textbf{Qwen3-32B} & \textbf{Qwen2.5-72B} & \textbf{LLaMA3.3-70B} \\ \midrule
Baseline & 29.00\textsubscript{±0.0} & 53.00\textsubscript{±2.9} & 80.50\textsubscript{±1.8} & 80.00\textsubscript{±2.6} & 73.00\textsubscript{±0.0} & 73.00\textsubscript{±0.0} & 51.50\textsubscript{±0.0} \\ \midrule

RAG & 36.00\textsubscript{±0.0} & 65.50\textsubscript{±1.7} & \textbf{85.50}\textsubscript{±1.8} & 81.50\textsubscript{±2.5} & \textbf{83.00}\textsubscript{±0.0} & 76.50\textsubscript{±0.0} & 70.50\textsubscript{±0.0} \\
RAG+ & \textbf{41.00}\textsubscript{±0.0} & \textbf{67.50}\textsubscript{±1.8} & \textbf{85.50}\textsubscript{±2.1} & \textbf{86.00}\textsubscript{±2.4} & 82.50\textsubscript{±0.0} & \textbf{83.00}\textsubscript{±0.0} & \textbf{76.00}\textsubscript{±0.0} \\ \midrule

AFRAG & 27.50\textsubscript{±0.0} & 65.50\textsubscript{±2.2} & 85.00\textsubscript{±2.6} & 82.50\textsubscript{±2.9} & 76.00\textsubscript{±0.0} & 85.00\textsubscript{±0.0} & 41.50\textsubscript{±0.0} \\
AFRAG+ & \textbf{33.00}\textsubscript{±0.0} & \textbf{68.00}\textsubscript{±2.3} & \textbf{86.50}\textsubscript{±2.5} & \textbf{83.00}\textsubscript{±2.9} & \textbf{77.50}\textsubscript{±0.0} & \textbf{86.50}\textsubscript{±0.0} & \textbf{52.50}\textsubscript{±0.0} \\ \midrule

GraphRAG & 36.50\textsubscript{±0.0} & 42.00\textsubscript{±2.8} & \textbf{81.50}\textsubscript{±2.2} & 76.00\textsubscript{±2.3} & 68.50\textsubscript{±0.0} & \textbf{64.00}\textsubscript{±0.0} & 38.50\textsubscript{±0.0} \\
GraphRAG+ & \textbf{46.00}\textsubscript{±0.0} & \textbf{47.50}\textsubscript{±1.6} & \textbf{81.50}\textsubscript{±2.5} & \textbf{77.50}\textsubscript{±2.7} & \textbf{75.00}\textsubscript{±0.0} & \textbf{64.00}\textsubscript{±0.0} & \textbf{52.00}\textsubscript{±0.0} \\ \midrule

Rerank RAG & 33.00\textsubscript{±0.0} & 60.00\textsubscript{±2.1} & 82.00\textsubscript{±1.8} & \textbf{83.50}\textsubscript{±2.0} & 80.50\textsubscript{±0.0} & 77.50\textsubscript{±0.0} & \textbf{77.50}\textsubscript{±0.0} \\
Rerank RAG+ & \textbf{34.00}\textsubscript{±0.0} & \textbf{61.00}\textsubscript{±1.1} & \textbf{82.50}\textsubscript{±1.9} & \textbf{83.50}\textsubscript{±2.1} & \textbf{82.00}\textsubscript{±0.0} & \textbf{87.50}\textsubscript{±0.0} & \textbf{77.50}\textsubscript{±0.0} \\
\bottomrule
\end{tabular}
}
\end{table*}

\begin{table*}[tp]
\centering
\small
\caption{Accuracy (\%) of different models on the MedQA dataset with and without application-level augmentation.}
\label{tab:med_main_result_full}
\resizebox{\textwidth}{!}{
\begin{tabular}{l|ccccccc}
\toprule
\textbf{Method} & \textbf{LLaMA3.1-8B} & \textbf{Qwen2.5-7B} & \textbf{DS-Qwen-7B} & \textbf{DS-Qwen-32B} & \textbf{QwQ-32B} & \textbf{Qwen2.5-72B} & \textbf{LLaMA3.3-70B} \\ \midrule
Baseline & 57.80\textsubscript{±0.0} & 41.80\textsubscript{±0.0} & 32.60\textsubscript{±1.2} & 80.00\textsubscript{±0.8} & 80.40\textsubscript{±1.2} & 73.80\textsubscript{±0.0} & 78.20\textsubscript{±0.0} \\ \midrule

RAG & 63.00\textsubscript{±0.0} & \textbf{59.20}\textsubscript{±0.0} & 34.60\textsubscript{±1.6} & 79.00\textsubscript{±1.0}  & 80.20\textsubscript{±0.9} & 75.00\textsubscript{±0.0} & 80.20\textsubscript{±0.0} \\
RAG+ & \textbf{63.60}\textsubscript{±0.0} & 57.60\textsubscript{±0.0} & \textbf{40.20}\textsubscript{±0.5} & \textbf{80.20}\textsubscript{±0.5} & \textbf{80.80}\textsubscript{±0.4} & \textbf{75.40}\textsubscript{±0.0} & \textbf{81.40}\textsubscript{±0.0} \\ \midrule

AFRAG & 56.40\textsubscript{±0.0} & 53.40\textsubscript{±0.0} & 32.20\textsubscript{±1.3} & 78.20\textsubscript{±0.9} & 81.20\textsubscript{±1.6} & 76.40\textsubscript{±0.0} & 82.40\textsubscript{±0.0} \\
AFRAG+ & \textbf{57.00}\textsubscript{±0.0} & \textbf{57.20}\textsubscript{±0.0} & \textbf{34.60}\textsubscript{±1.2} & \textbf{78.60}\textsubscript{±0.9} & \textbf{82.20}\textsubscript{±1.2} & \textbf{77.40}\textsubscript{±0.0} & \textbf{83.00}\textsubscript{±0.0} \\ \midrule

Rerank RAG & 60.00\textsubscript{±0.0} & 58.60\textsubscript{±0.0} & \textbf{35.20}\textsubscript{±0.8} & 79.80\textsubscript{±0.5} & 80.60\textsubscript{±0.9} & 76.40\textsubscript{±0.0} & 81.00\textsubscript{±0.0} \\
Rerank RAG+ & \textbf{63.40}\textsubscript{±0.0} & \textbf{61.40}\textsubscript{±0.0} & -- & \textbf{80.20}\textsubscript{±1.0} & \textbf{81.40}\textsubscript{±0.7} & \textbf{78.20}\textsubscript{±0.0} & \textbf{85.60}\textsubscript{±0.0} \\
\bottomrule
\end{tabular}
}
\end{table*}

\begin{table*}[t]
\centering
\small
\caption{Accuracy (\%) of different models on the Math dataset with and without application-level augmentation.}
\label{tab:na_main_results_full}
\resizebox{\textwidth}{!}{
\begin{tabular}{l|ccccccc}
\toprule
\textbf{Method} & \textbf{GLM4-9B} & \textbf{Qwen2.5-7B} & \textbf{DS-Qwen-7B} & \textbf{Qwen2.5-14B} & \textbf{DS-Qwen-32B} & \textbf{Qwen2.5-72B} & \textbf{LLaMA3.3-70B} \\ 
\midrule

Baseline & 46.51\textsubscript{±0.0} & 58.84\textsubscript{±0.0} & 24.19\textsubscript{±0.7} & 66.98\textsubscript{±0.0} & 80.00\textsubscript{±0.6} & 69.07\textsubscript{±0.0} & 69.07\textsubscript{±0.0} \\ 
\midrule

RAG & 47.21\textsubscript{±0.0} & 64.65\textsubscript{±0.0} & 24.42\textsubscript{±1.0} & 73.49\textsubscript{±0.0} & 82.79\textsubscript{±1.2} & 70.47\textsubscript{±0.0} & 71.16\textsubscript{±0.0} \\
RAG+ & \textbf{52.09}\textsubscript{±0.0} & \textbf{65.58}\textsubscript{±0.0} & \textbf{26.28}\textsubscript{±0.7} & \textbf{74.67}\textsubscript{±0.0} & \textbf{84.65}\textsubscript{±1.3} & \textbf{73.72}\textsubscript{±0.0} & \textbf{71.86}\textsubscript{±0.0} \\ 
\midrule

AFRAG & 48.14\textsubscript{±0.0} & 63.51\textsubscript{±0.0} & 22.56\textsubscript{±0.6} & 66.98\textsubscript{±0.0} & 82.09\textsubscript{±0.5} & 71.86\textsubscript{±0.0} & 70.23\textsubscript{±0.0} \\
AFRAG+ & \textbf{51.16}\textsubscript{±0.0} & \textbf{64.42}\textsubscript{±0.0} & \textbf{23.95}\textsubscript{±0.8} & \textbf{70.00}\textsubscript{±0.0} & \textbf{83.95}\textsubscript{±0.6} & \textbf{76.05}\textsubscript{±0.0} & \textbf{71.86}\textsubscript{±0.0} \\ 
\midrule

GraphRAG & 33.95\textsubscript{±0.0} & 56.98\textsubscript{±0.0} & 27.21\textsubscript{±0.8} & 69.07\textsubscript{±0.0} & 82.79\textsubscript{±0.6} & \textbf{73.02}\textsubscript{±0.0} & 68.37\textsubscript{±0.0} \\
GraphRAG+ & \textbf{36.51}\textsubscript{±0.0} & \textbf{59.77}\textsubscript{±0.0} & \textbf{33.72}\textsubscript{±0.8} & \textbf{69.77}\textsubscript{±0.0} & \textbf{83.49}\textsubscript{±0.5} & 72.56\textsubscript{±0.0} & \textbf{69.00}\textsubscript{±0.0} \\ 
\midrule

Rerank RAG & \textbf{48.21}\textsubscript{±0.0} & 56.05\textsubscript{±0.0} & 26.05\textsubscript{±0.9} & 71.40\textsubscript{±0.0} & 80.46\textsubscript{±0.7} & 73.26\textsubscript{±0.0} & 74.65\textsubscript{±0.0} \\
Rerank RAG+ & -- & \textbf{56.28}\textsubscript{±0.0} & \textbf{32.09}\textsubscript{±0.6} & \textbf{78.90}\textsubscript{±0.0} & \textbf{83.26}\textsubscript{±0.5} & \textbf{77.21}\textsubscript{±0.0} & \textbf{76.74}\textsubscript{±0.0} \\
\bottomrule
\end{tabular}
}
\end{table*}

Based on the results in Table~\ref{tab:med-rag-}, incorporating application examples alone improves performance over the base models in several cases, particularly for Qwen2.5-14B and Qwen2.5-72B. However, combining both knowledge and application examples generally yields better results, as shown by the superior accuracy of RAG+ and AFRAG+ across most models.
\begin{table*}[htp]
\small
\centering
\caption{Performance comparison of application-only retrieval vs knowledge-enhanced retrieval on the MedQA dataset.}
\label{tab:med-rag-}
\begin{tabular}{c|ccccc}
\toprule
\textbf{Method} &  \textbf{Qwen2.5-14B} & \textbf{Qwen3-32B} & \textbf{DS-Qwen-32B} & \textbf{Qwen2.5-72B} \\ \midrule
Baseline &  68.80 & 84.40 & 79.00 & 73.80 \\ \midrule
RAG &  68.40 & 82.60 & 81.00 & 75.00 \\
RAG-app & \textbf{70.60} & 83.60 & 78.40 & \textbf{76.60} \\
RAG-plus  & 69.80 & \textbf{84.20} & \textbf{82.20} & 75.20 \\ \midrule
AFRAG  & 69.00 & 83.00 & 80.20 & 76.40 \\
AFRAG-app  & 70.60 & 82.42 & 77.20 & \textbf{77.80} \\
AFRAG+  & 69.40 & \textbf{83.60} & \textbf{80.60} & 77.40 \\ \bottomrule
\end{tabular}
\end{table*}

\subsection{Model Scale}
Table~\ref{apptab:model_scale_legal} presents the performance of Qwen2.5 models of different scales on the sentencing prediction task in the legal domain, and Table~\ref{apptab:model_scale_med} shows their performance on the MedQA dataset.
\begin{table*}[htp]
\small
\centering
\caption{Accuracy (\%) Comparison of Qwen2.5 Models of Varying Scales on Sentencing Prediction.}\label{apptab:model_scale_legal}
\begin{tabularx}{0.9\textwidth}{C|CCC}
\toprule
\textbf{Method} & \textbf{Qwen2.5-7B} & \textbf{Qwen2.5-14B} & \textbf{Qwen2.5-72B} \\
 \midrule
 
Baseline & 46.00 & 74.00 &  58.00 \\  \midrule

RAG & 75.00 & 78.50 & 76.50 \\
RAG+ & \textbf{76.50} & \textbf{82.50} & \textbf{83.00} \\  \midrule

AF-RAG & 40.50 & 81.00 &  85.00 \\
AF-RAG+ & \textbf{47.00} & \textbf{85.50} & \textbf{85.50} \\  \midrule

GraphRAG & 47.00 & 68.00 &  64.00 \\
GraphRAG+ & \textbf{59.00} & \textbf{79.00} &  \textbf{64.00} \\  \midrule

Rerank RAG & 70.50 & 78.50 &  77.50 \\
Rerank RAG+ & \textbf{74.50} & \textbf{81.00} & \textbf{87.50} \\ \bottomrule
\end{tabularx}
\end{table*}
\begin{table*}[htp]
\small
\centering
\caption{Accuracy (\%) Comparison of Qwen2.5 Models of Varying Scales on the MedQA Dataset.}\label{apptab:model_scale_med}
\begin{tabularx}{0.9\textwidth}{C|CCC}
\toprule
 \textbf{Method} & \textbf{Qwen2.5-7B} & \textbf{Qwen2.5-14B} &  \textbf{Qwen2.5-72B} \\
 \midrule
Baseline & 41.80 & 68.80 & 73.80 \\  \midrule

RAG & \textbf{59.20} & 68.40 &  75.00 \\
RAG+ & 57.60 & \textbf{69.80} &  \textbf{75.20} \\  \midrule

AF-RAG & 53.40 & 69.00 &  76.40 \\
AF-RAG+ & \textbf{57.20} & \textbf{69.40} &  \textbf{77.40} \\  \midrule

Rerank RAG & 58.60 & 63.80 &  76.40 \\
Rerank RAG+ & \textbf{61.40} & \textbf{68.40} &  \textbf{78.20} \\
\bottomrule
\end{tabularx}
\end{table*}

\subsection{Case Study}
\begin{figure*}[t]
\centering
\includegraphics[width=\textwidth]{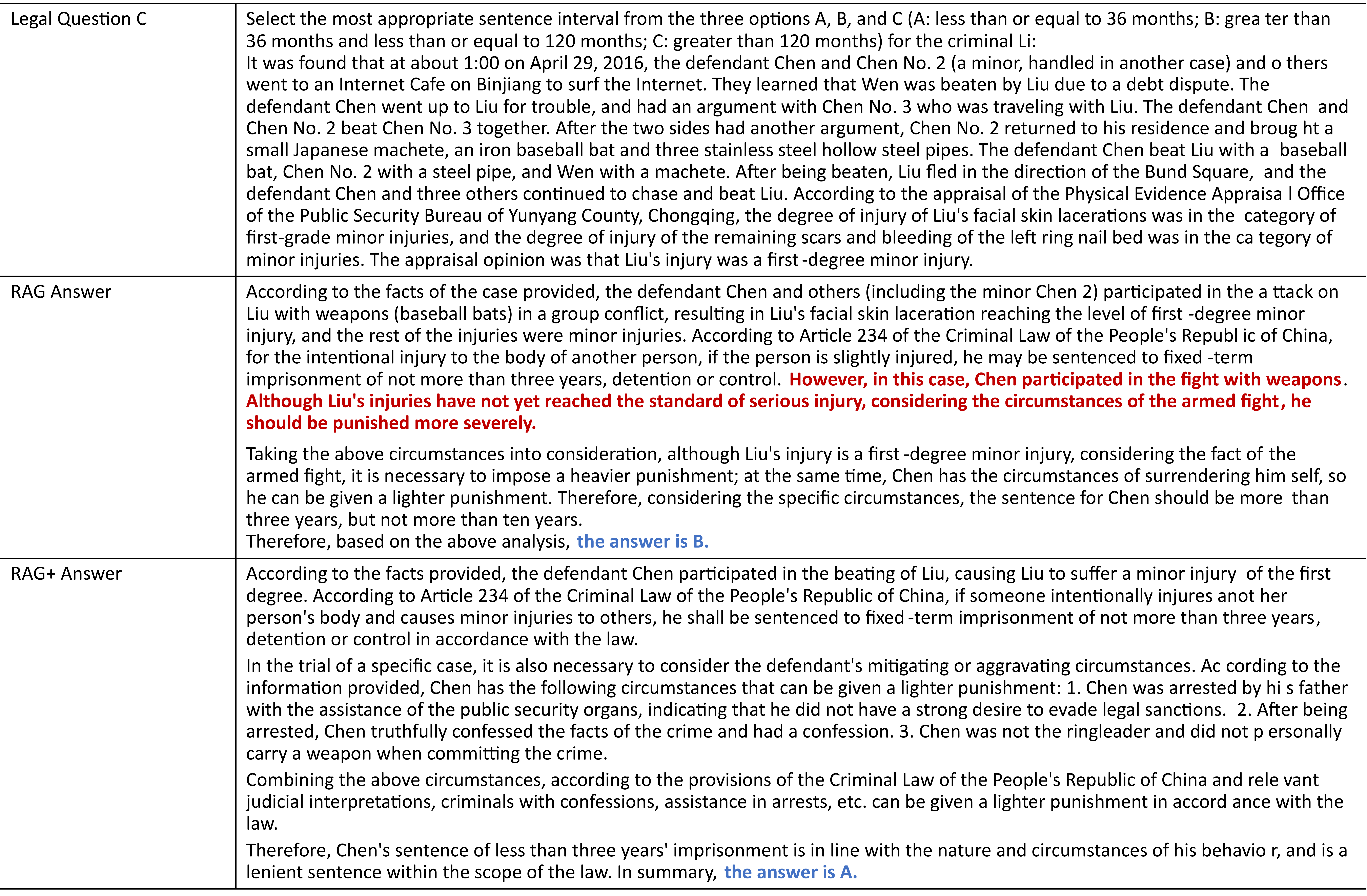}
\caption{Comparison of Qwen3-32B Outputs under RAG and RAG+ Configurations on Sentencing Prediction.}
\label{appendfig:casestudylegal}
\end{figure*}

\begin{figure*}[t]
\centering
\includegraphics[width=\textwidth]{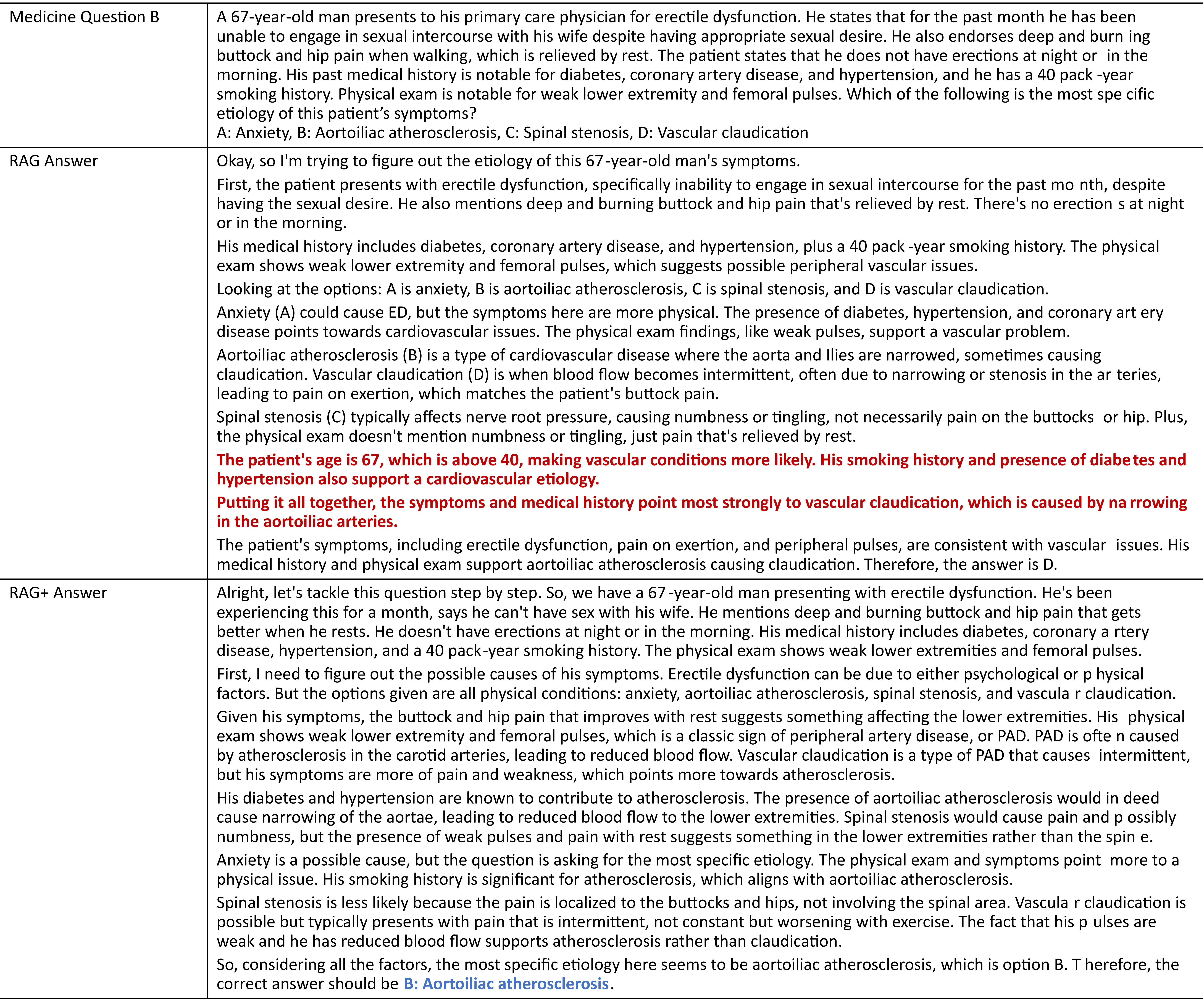}
\caption{Comparison of DeepSeek-R1-Distill-Qwen-7B model Outputs under RAG and RAG+ Configurations on MedQA.}
\label{appendfig:casestudymedicine}
\end{figure*}

We present two case studies in Figures~\ref{appendfig:casestudylegal} and~\ref{appendfig:casestudymedicine}, comparing the outputs of Qwen3-32B and DeepSeek-R1-Distill-Qwen-7B under RAG and RAG+ configurations on the sentencing prediction and MedQA dataset, respectively. The results demonstrate that merely retrieving external knowledge, as done in standard RAG, is often insufficient: models may still make reasoning mistakes or misuse the retrieved content. In contrast, with the integration of the application module in RAG+, the models are able to apply the retrieved knowledge more appropriately, leading to correct predictions. These qualitative results further support the effectiveness of RAG+ in enhancing knowledge utilization during inference.

\section{Prompts}\label{app:prompt}
\begin{figure*}[t]
\centering
\includegraphics[width=0.95\textwidth]{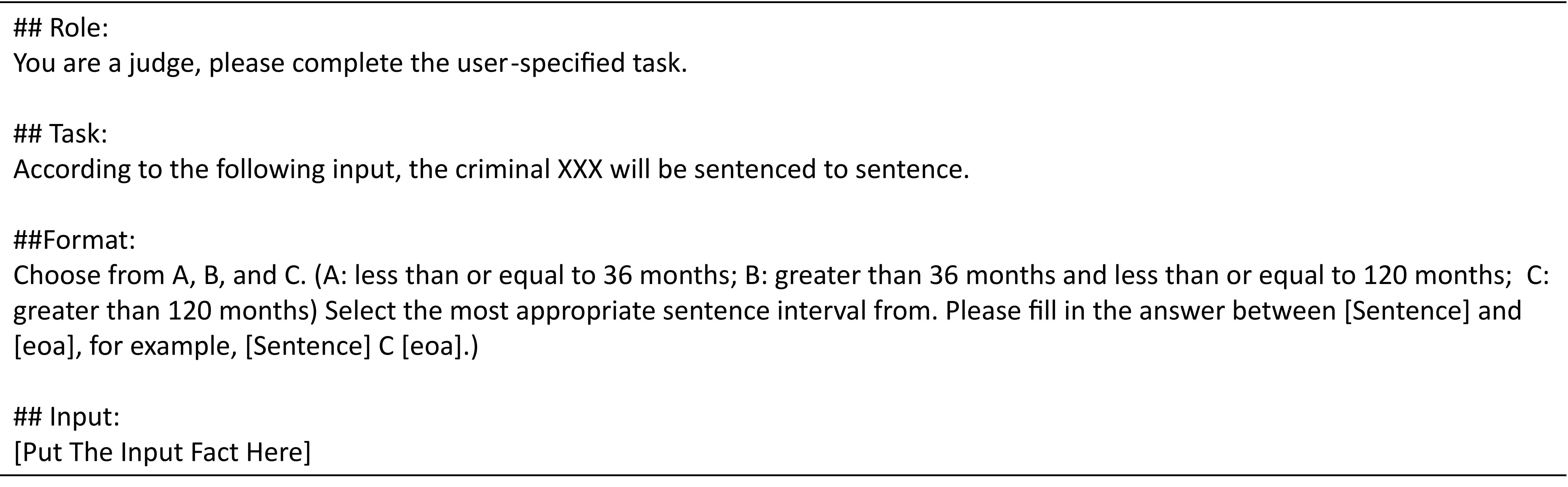}
\caption{An Example of the Prompt Template Used for the Sentencing Prediction Task under the Base Configuration.}
\label{appfig:prompt-legal-base}
\end{figure*}

\begin{figure*}[t]
\centering
\includegraphics[width=0.95\textwidth]{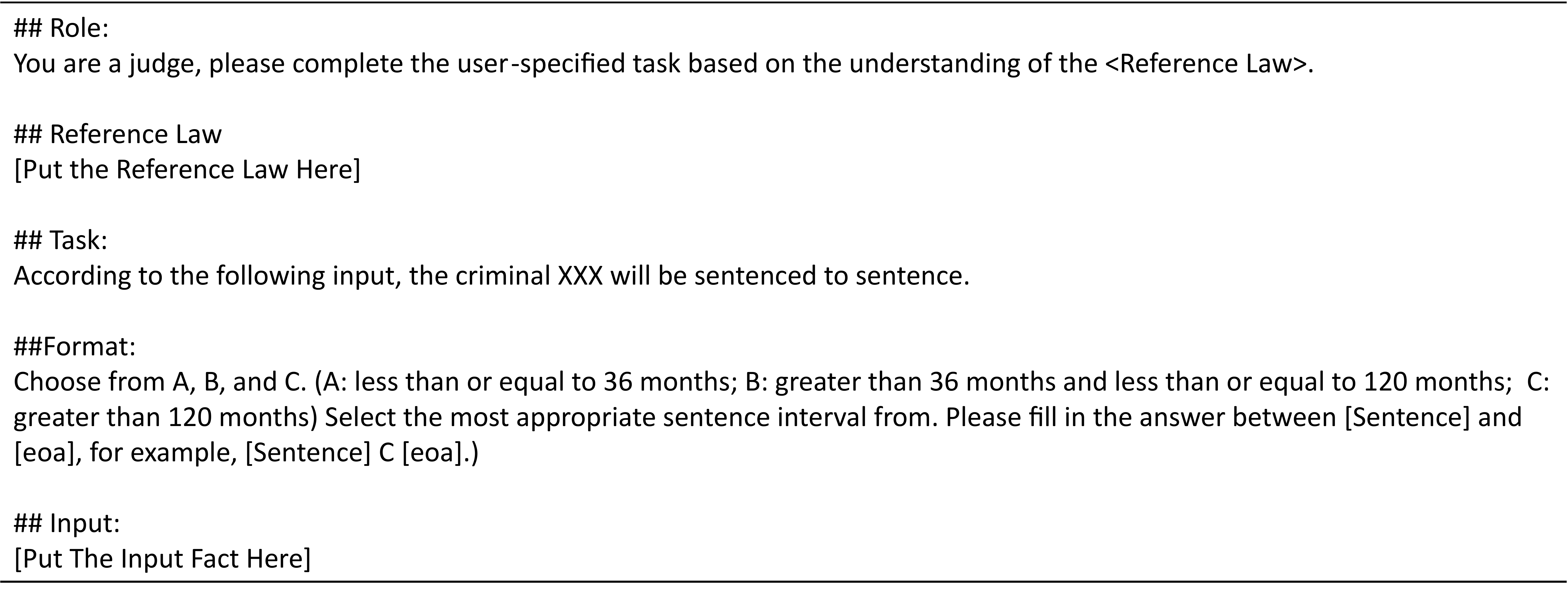}
\caption{An Example of the Prompt Template Used for the Sentencing Prediction Task under the RAG Configuration.}
\label{appfig:prompt-legal-RAG}
\end{figure*}

\begin{figure*}[t]
\centering
\includegraphics[width=0.95\textwidth]{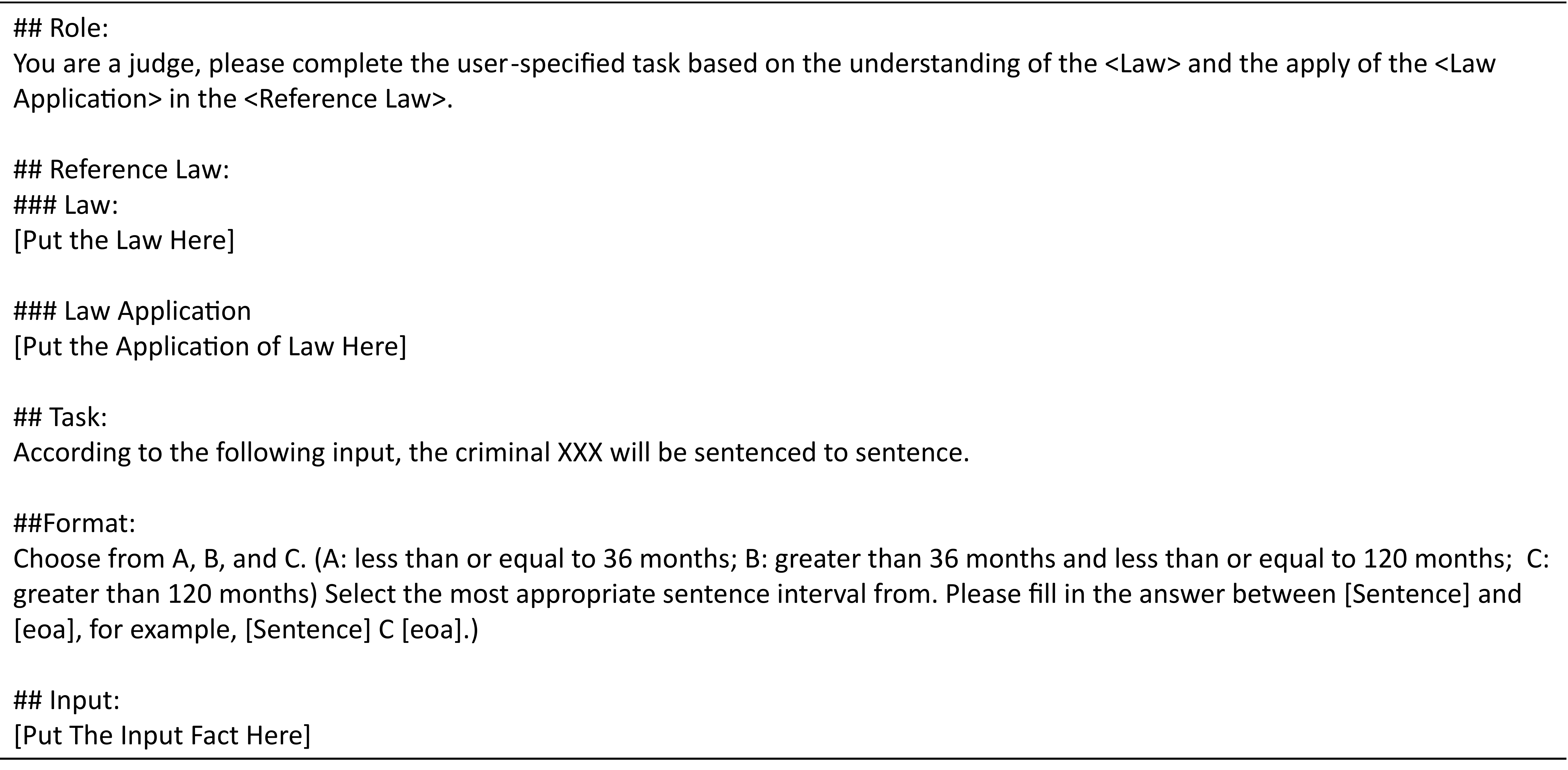}
\caption{An Example of the Prompt Template Used for the Sentencing Prediction Task under the RAG+ Configuration.}
\label{appfig:prompt-legal-RAG+}
\end{figure*}

\begin{figure*}[t]
\centering
\includegraphics[width=0.95\textwidth]{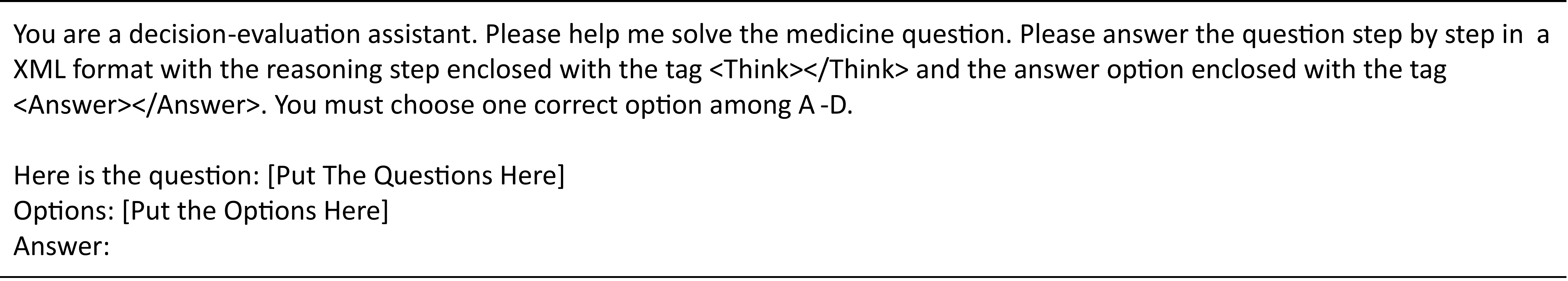}
\caption{An Example of the Prompt Template Used for the MedQA Dataset under the Base Configuration.}
\label{appfig:prompt-med-base}
\end{figure*}

\begin{figure*}[t]
\centering
\includegraphics[width=0.95\textwidth]{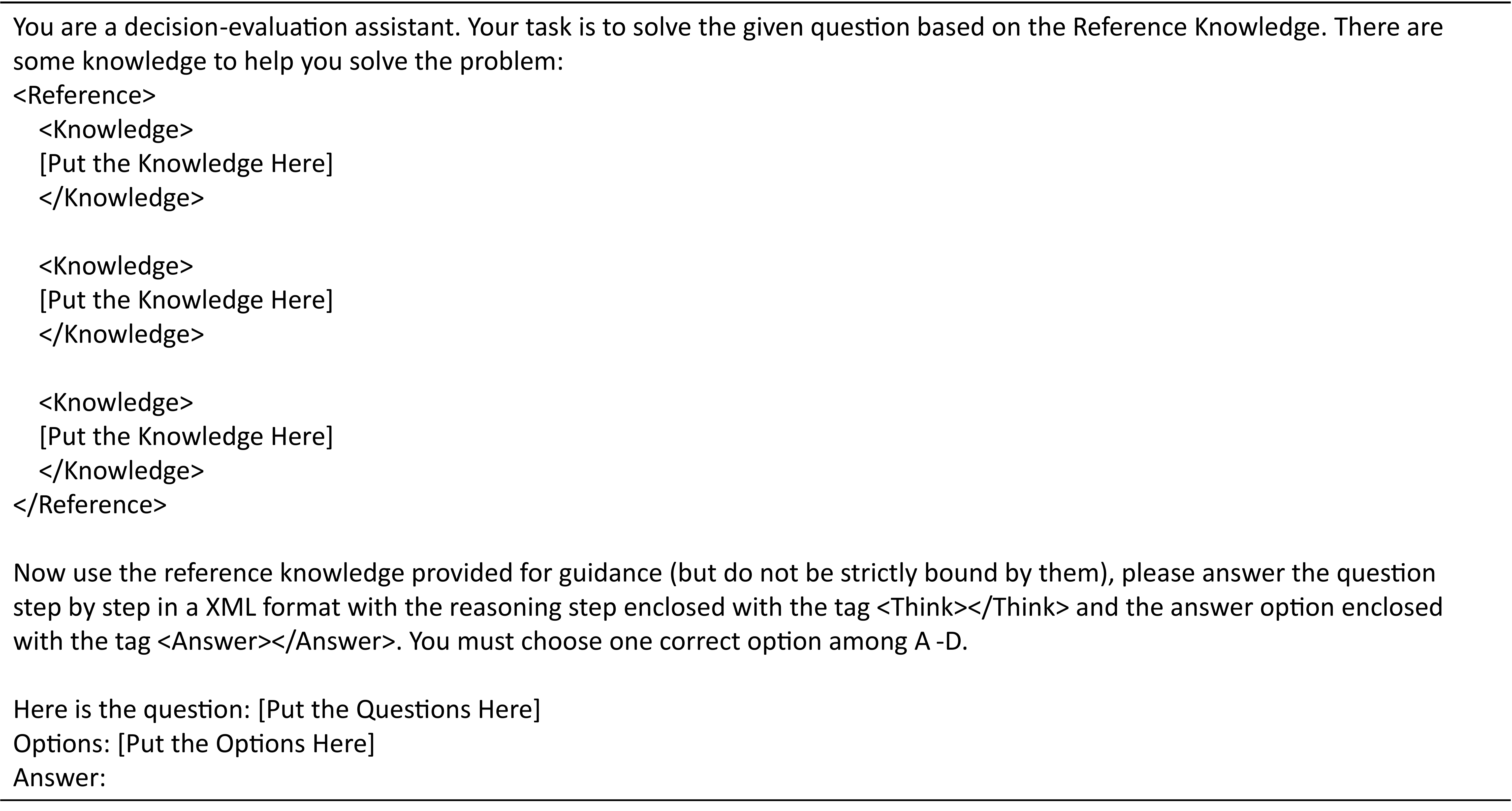}
\caption{An Example of the Prompt Template Used for the MedQA Dataset under the RAG Configuration.}
\label{appfig:prompt-med-RAG}
\end{figure*}

\begin{figure*}[t]
\centering
\includegraphics[width=0.95\textwidth]{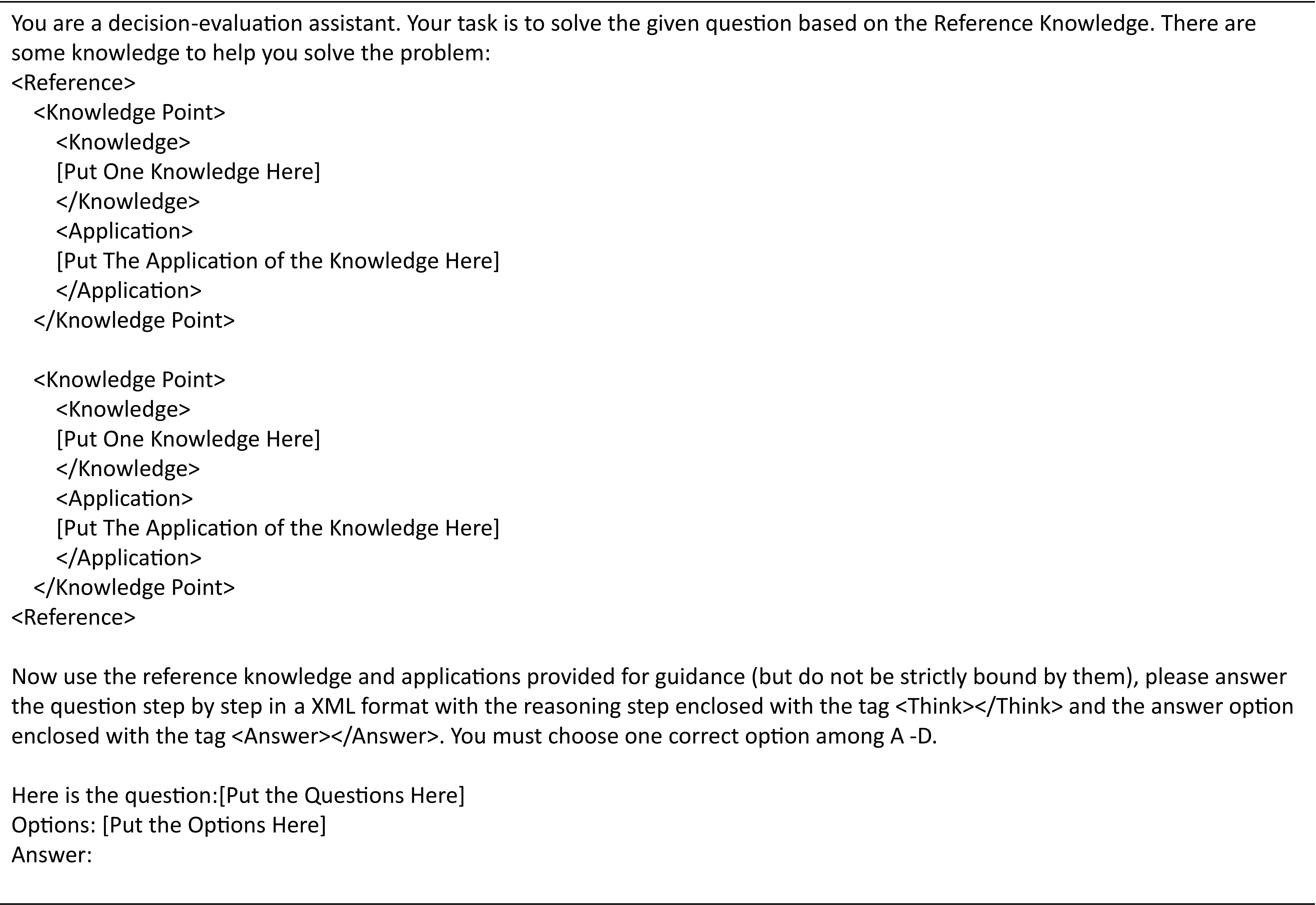}
\caption{An Example of the Prompt Template Used for the MedQA Dataset under the RAG+ Configuration.}
\label{appfig:prompt-med-RAG+}
\end{figure*}

\begin{figure*}[t]
\centering
\includegraphics[width=0.95\textwidth]{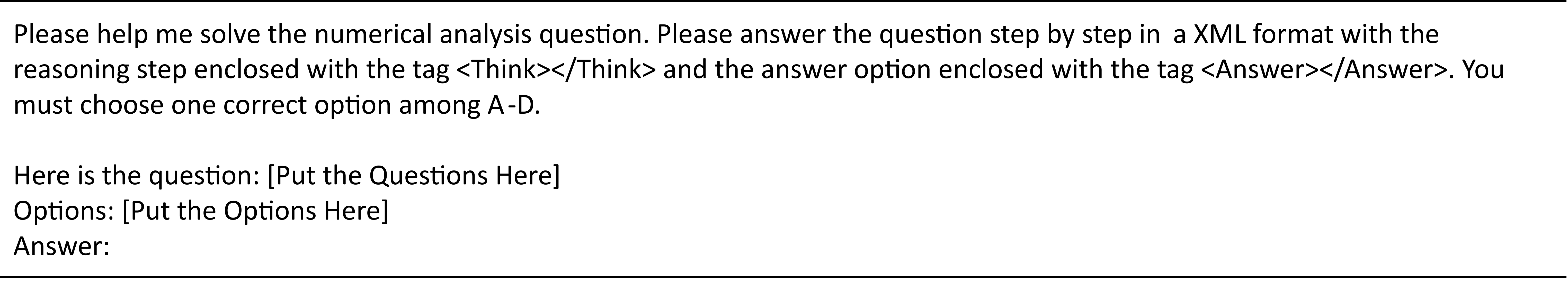}
\caption{An Example of the Prompt Template Used for the Math Task under the Base Configuration.}
\label{appfig:prompt-math-base}
\end{figure*}

\begin{figure*}[t]
\centering
\includegraphics[width=0.95\textwidth]{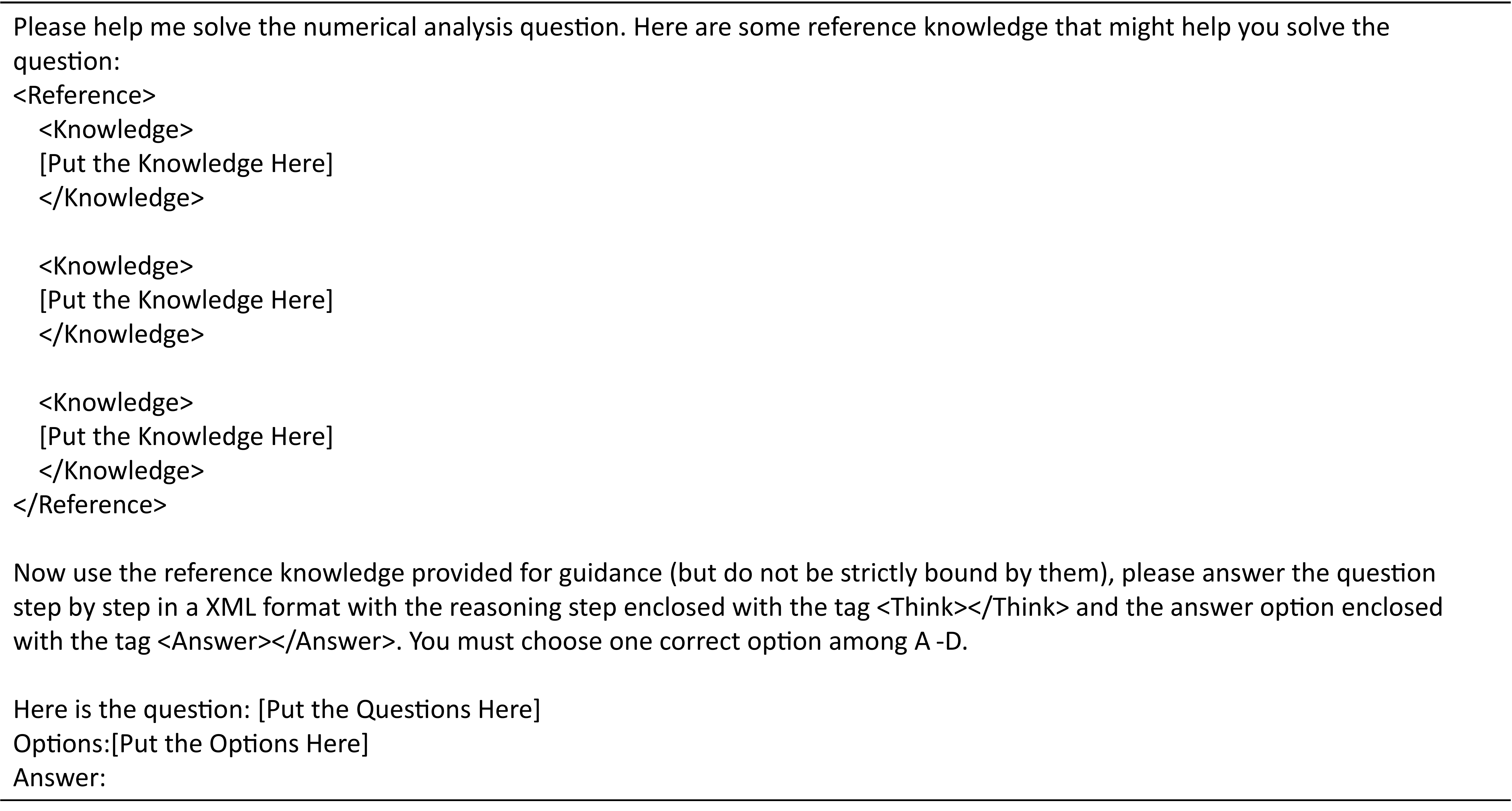}
\caption{An Example of the Prompt Template Used for the Math Task under the RAG Configuration.}
\label{appfig:prompt-math-RAG}
\end{figure*}

\begin{figure*}[t]
\centering
\includegraphics[width=0.95\textwidth]{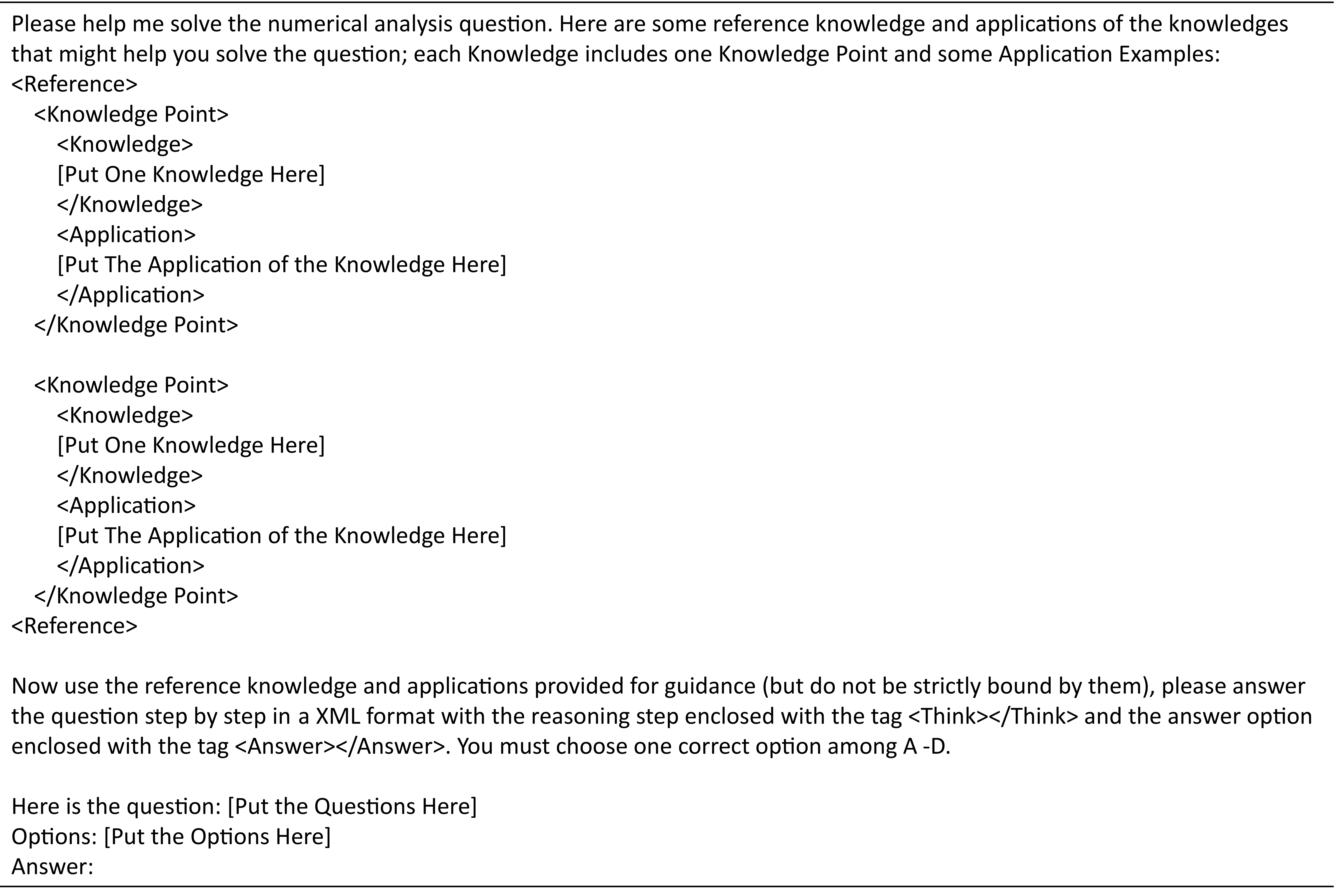}
\caption{An Example of the Prompt Template Used for the Math Task under the RAG+ Configuration.}
\label{appfig:prompt-math-RAG+}
\end{figure*}

The prompt configurations used across three domains—legal, medical, and mathematical—under the Base, RAG, and RAG+ settings are shown in Figures~\ref{appfig:prompt-legal-base} to~\ref{appfig:prompt-math-RAG+}. Specifically, Figures~\ref{appfig:prompt-legal-base},~\ref{appfig:prompt-legal-RAG}, and~\ref{appfig:prompt-legal-RAG+} illustrate the prompt designs for the legal domain, while similar configurations for the medical, and mathematical domains are provided in Figures~\ref{appfig:prompt-med-base} to\ref{appfig:prompt-med-RAG+}, and~\ref{appfig:prompt-math-base} to~\ref{appfig:prompt-math-RAG+}, respectively. These templates clearly reflect how retrieved knowledge is introduced and applied in each configuration, enabling consistent comparison across domains and setups.

\subsection{Dataset}\label{sec:app-prompts-datasets}
\begin{figure*}[t]
\centering
\includegraphics[width=0.95\textwidth]{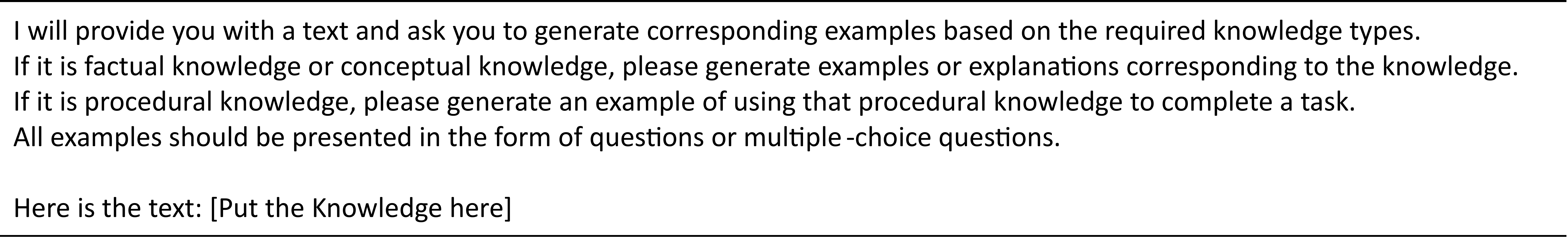}
\caption{The prompt template of generating the applications for the knowledge in legal domain.}
\label{appfig:legal-knowledge-generation}
\end{figure*}

\begin{figure*}[t]
\centering
\includegraphics[width=0.95\textwidth]{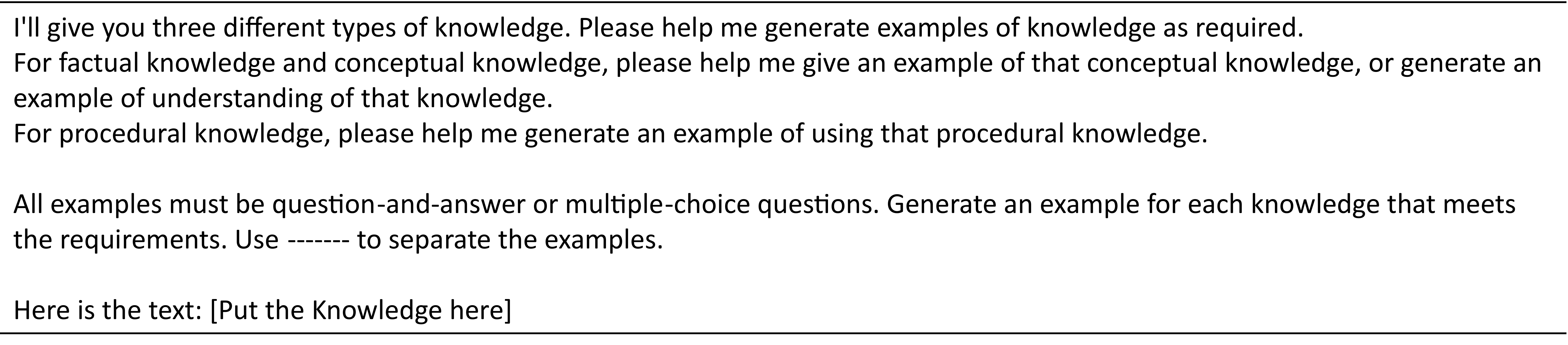}
\caption{The prompt template of generating the applications for the knowledge in medicine domain.}
\label{appfig:med-knowledge-generation}
\end{figure*}

\begin{figure*}[t]
\centering
\includegraphics[width=0.95\textwidth]{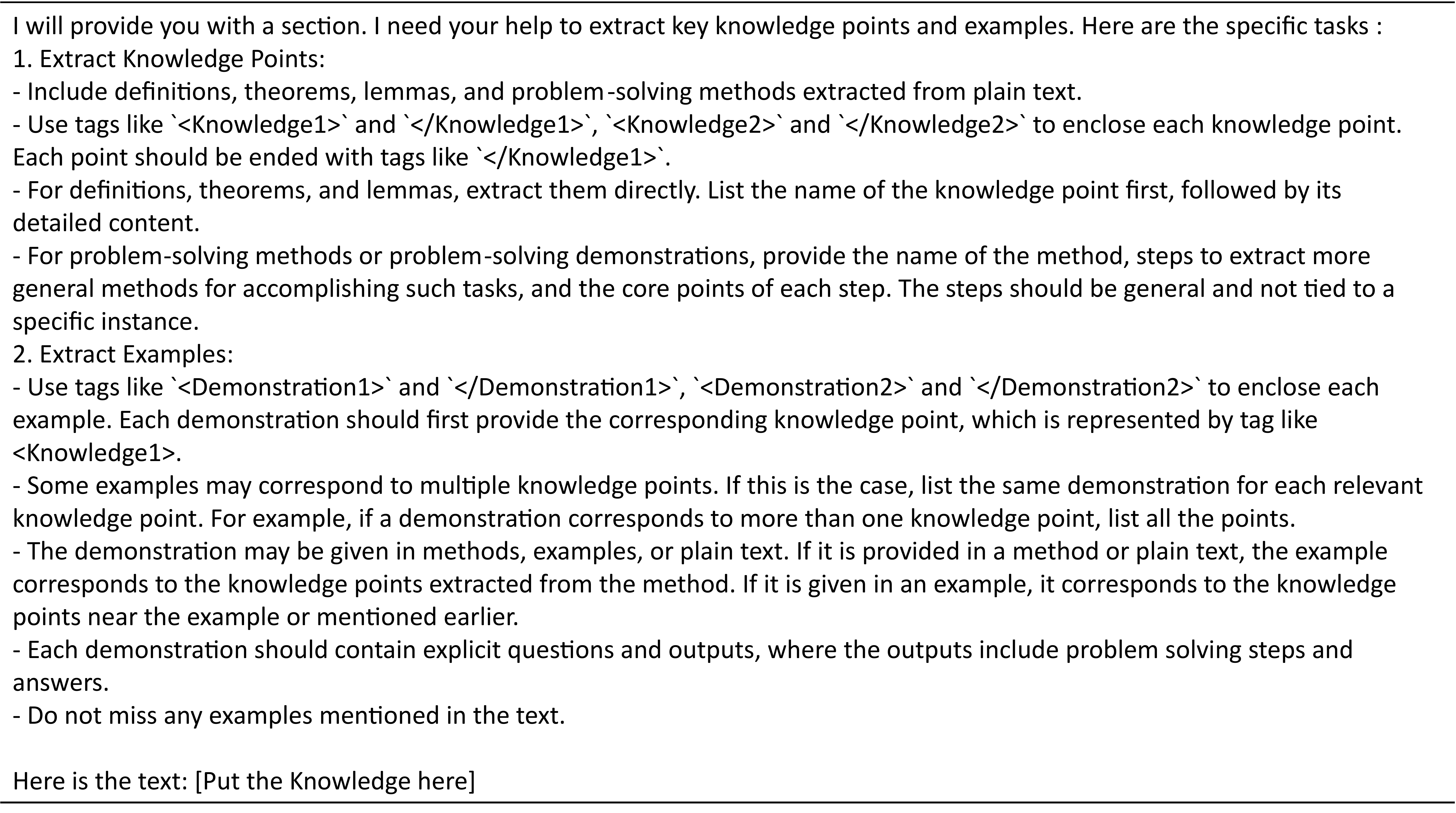}
\caption{The prompt template of generating the applications for the knowledge in mathematic domain.}
\label{appfig:math-knowledge-generation}
\end{figure*}

The prompt templates used to generate application examples for different domains are shown in Figures~\ref{appfig:legal-knowledge-generation},~\ref{appfig:med-knowledge-generation}, and~\ref{appfig:math-knowledge-generation}, corresponding to the legal, medical, and mathematical domains, respectively. These templates are designed to guide the model in producing domain-specific knowledge applications that align with the downstream tasks.

\subsection{Examples of Knowledge and Applications}
\begin{figure*}[t]
\centering
\includegraphics[width=0.95\textwidth]{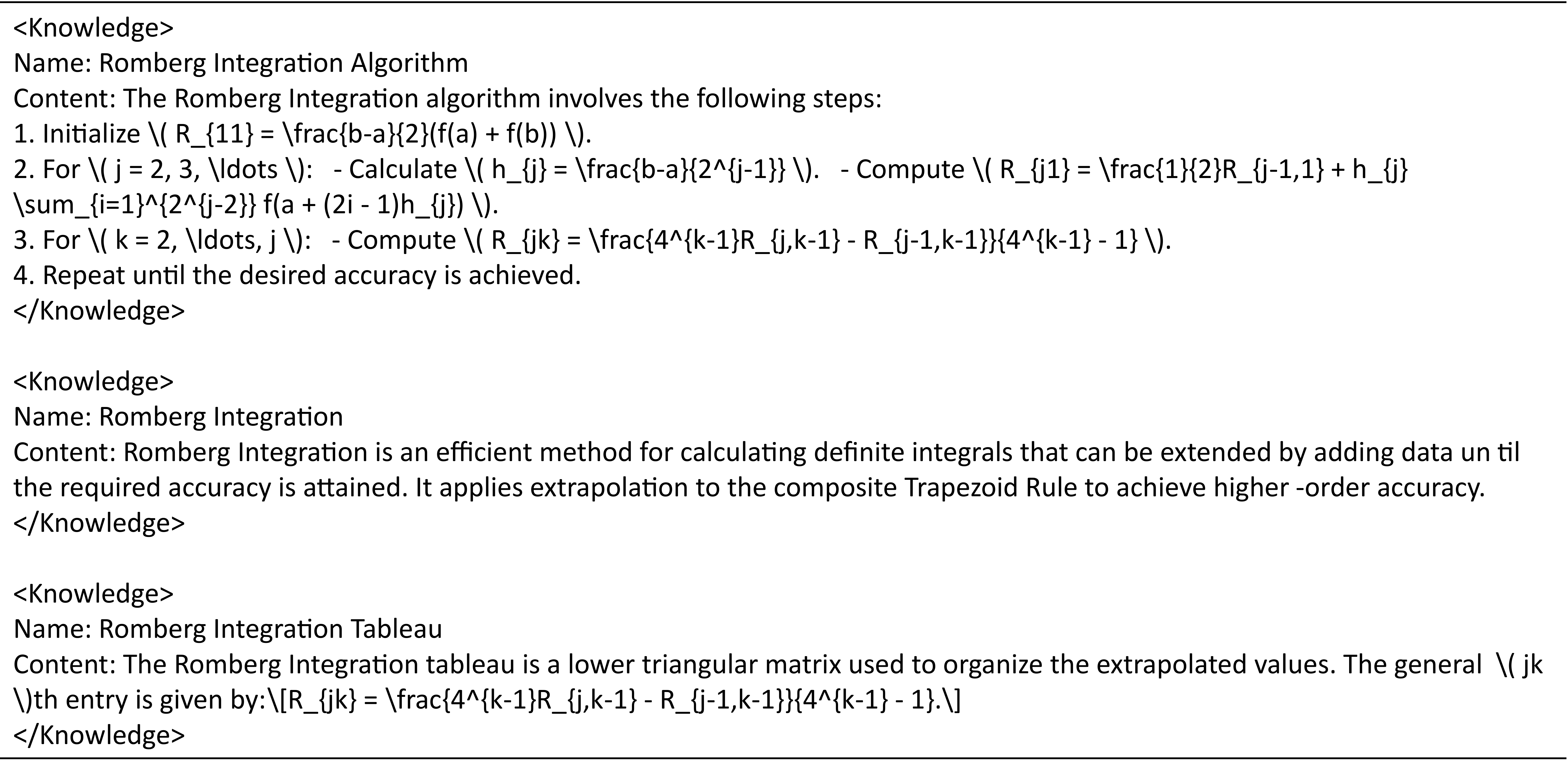}
\caption{Examples of the knowledge in Mathematic Domain.}
\label{appfig:math-knowledge-rag}
\end{figure*}

\begin{figure*}[t]
\centering
\includegraphics[width=0.95\textwidth]{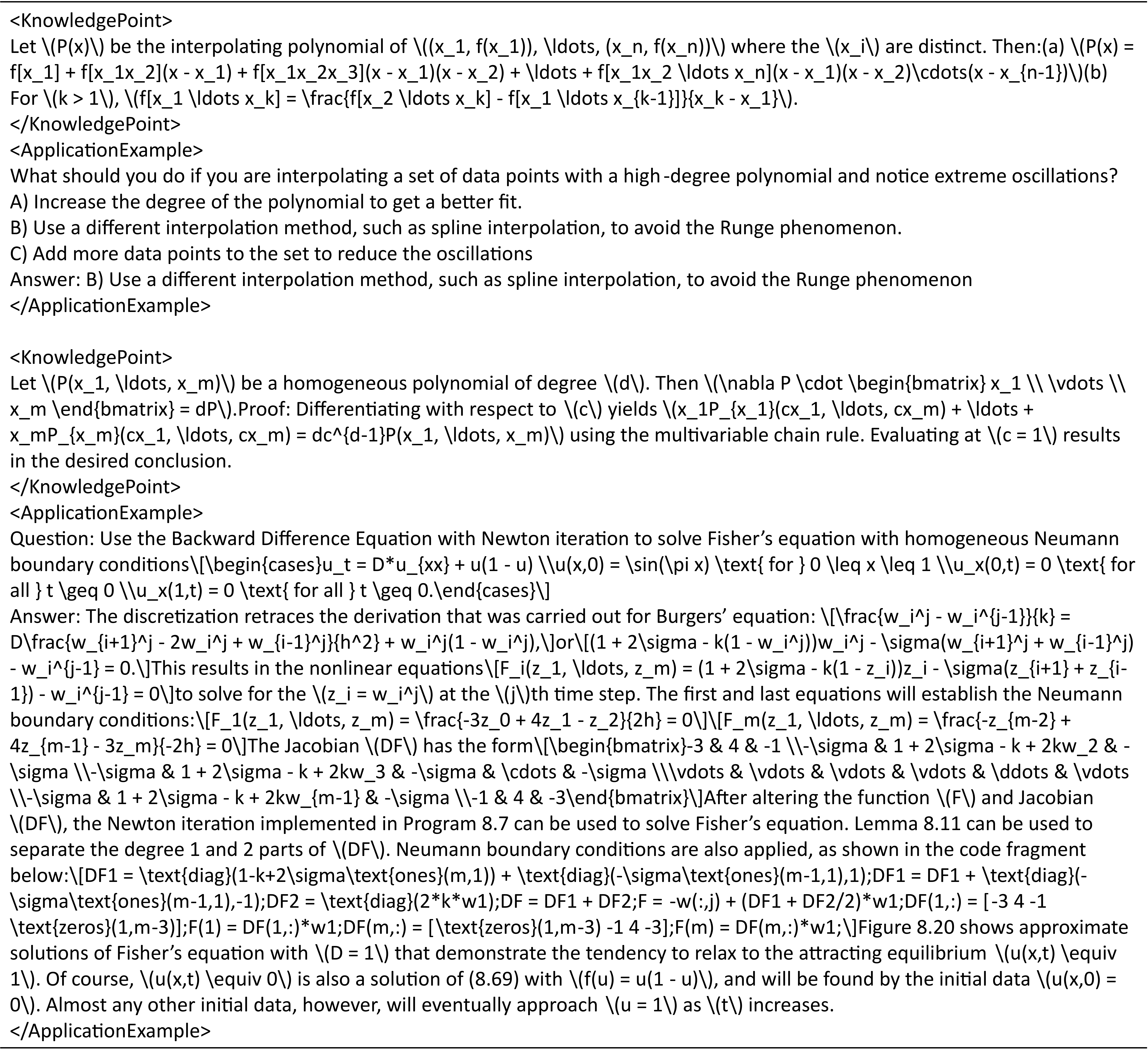}
\caption{Examples of the knowledge and the applications in Mathematic Domain.}
\label{appfig:math-knowledge-rag+}
\end{figure*}

\begin{figure*}[t]
\centering
\includegraphics[width=0.95\textwidth]{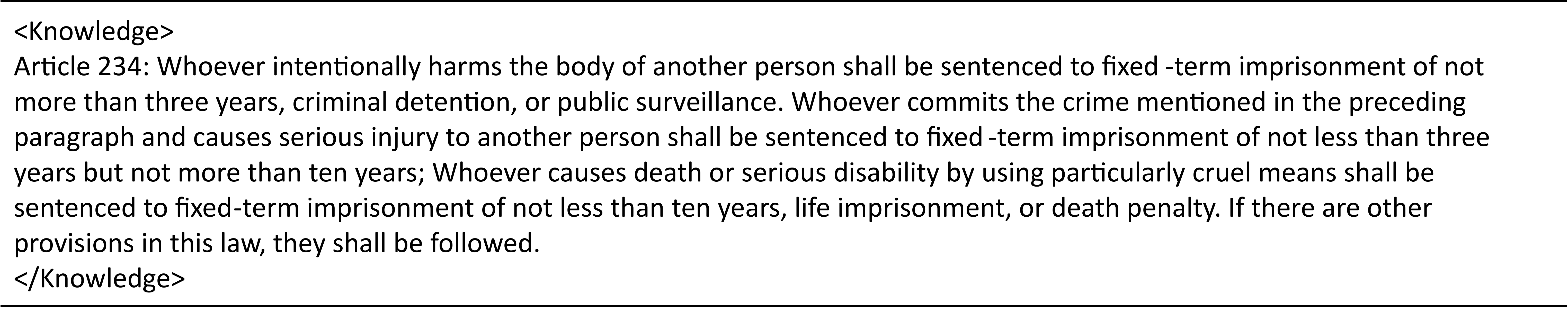}
\caption{Examples of the knowledge in legal Domain.}
\label{appfig:legal-knowledge-rag}
\end{figure*}

\begin{figure*}[t]
\centering
\includegraphics[width=0.95\textwidth]{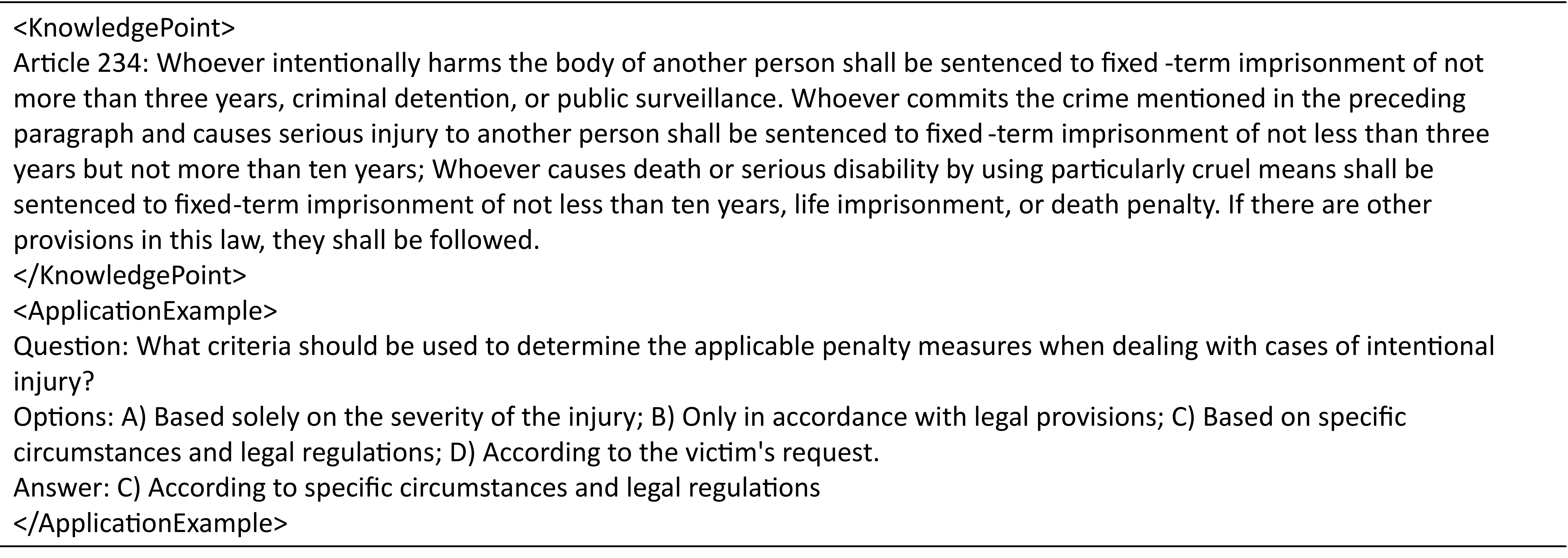}
\caption{Examples of the knowledge and the applications in legal Domain.}
\label{appfig:legal-knowledge-rag+}
\end{figure*}

\begin{figure*}[t]
\centering
\includegraphics[width=0.95\textwidth]{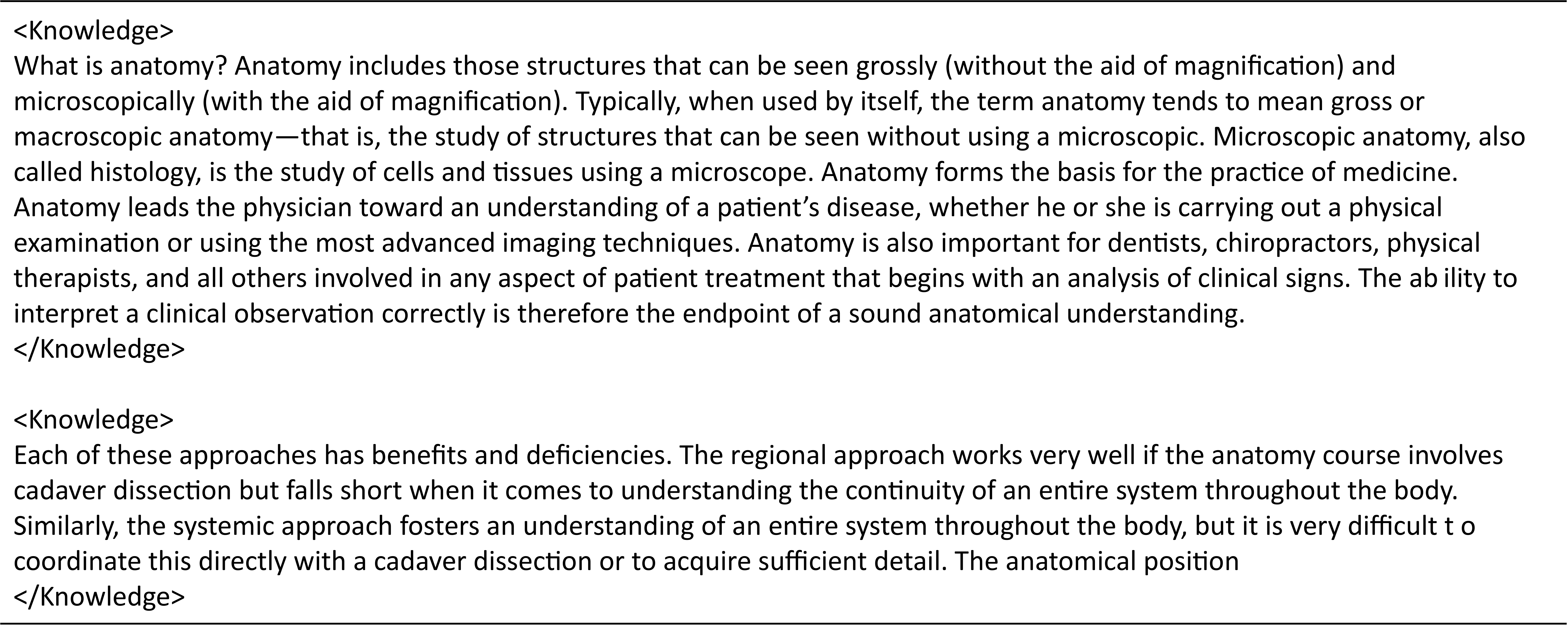}
\caption{Examples of the knowledge in Medicine Domain.}
\label{appfig:med-knowledge-rag}
\end{figure*}

\begin{figure*}[t]
\centering
\includegraphics[width=0.95\textwidth]{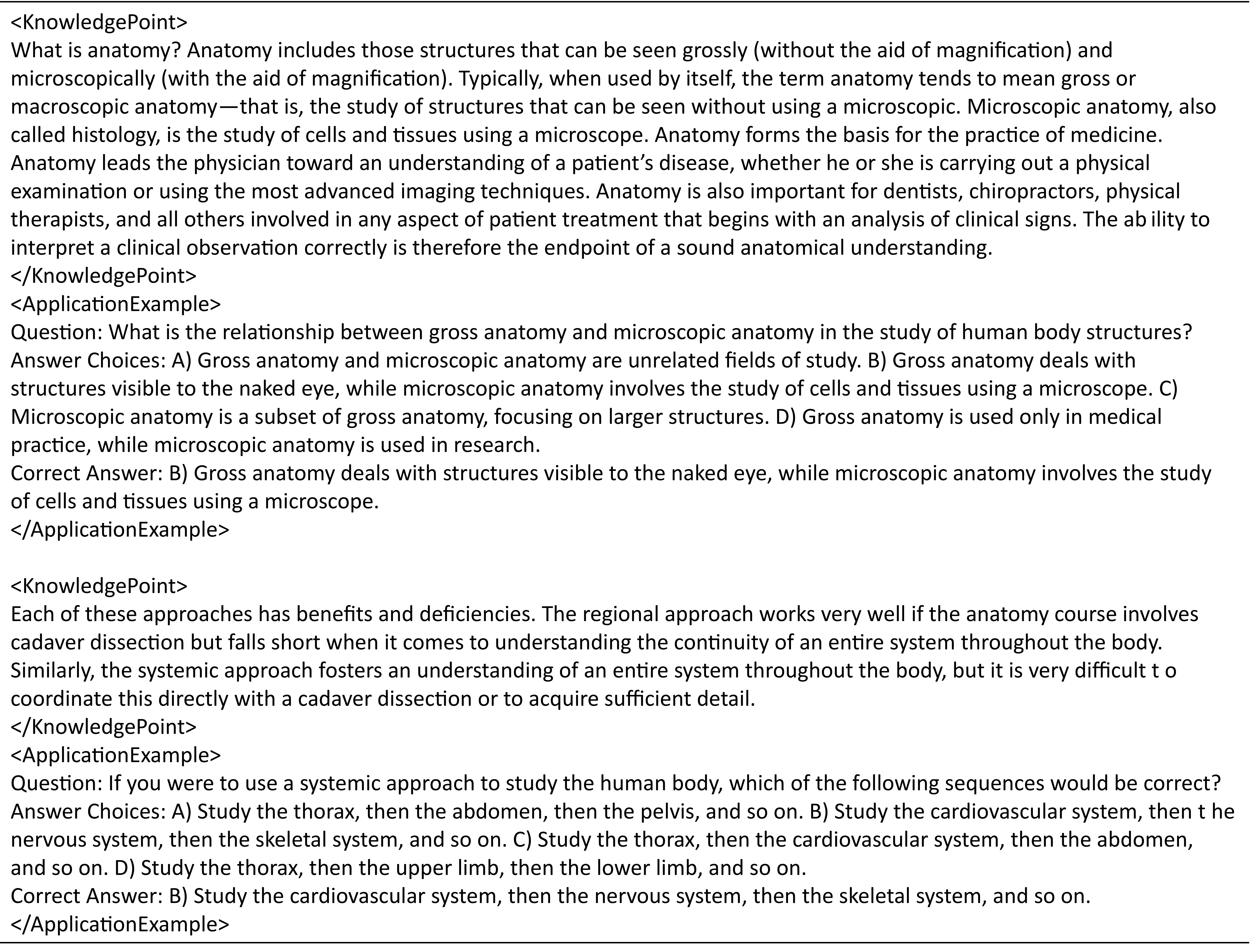}
\caption{Examples of the knowledge and the applications in Medicine Domain.}
\label{appfig:med-knowledge-rag+}
\end{figure*}

Examples of the retrieved and applied knowledge in different domains are shown in Figures~\ref{appfig:math-knowledge-rag} and~\ref{appfig:math-knowledge-rag+} for the mathematical domain, Figures~\ref{appfig:legal-knowledge-rag} and~\ref{appfig:legal-knowledge-rag+} for the legal domain, and Figures~\ref{appfig:med-knowledge-rag} and~\ref{appfig:med-knowledge-rag+} for the medical domain. Each pair of figures illustrates the difference between knowledge produced under the RAG and RAG+ configurations, highlighting how RAG+ promotes more targeted and applicable knowledge generation in support of downstream reasoning.

\end{document}